\documentclass{article}

\usepackage[margin=1.25in]{geometry}
\usepackage[utf8]{inputenc}
\usepackage[T1]{fontenc}
\usepackage{amsfonts}
\usepackage{amssymb}
\usepackage{amsmath}
\usepackage{mathtools}
\usepackage{booktabs}
\usepackage{tabularx}
\usepackage{graphicx}
\usepackage{microtype}
\usepackage{multirow}
\usepackage{nicefrac}
\usepackage{enumitem}
\usepackage{float}
\usepackage{parskip}
\usepackage{kpfonts}
\usepackage[small]{caption}
\usepackage{subcaption}
\usepackage{url}
\usepackage[table]{xcolor}
\usepackage{xspace}
\usepackage[numbers,sort&compress]{natbib}

\definecolor{LinkBlue}{rgb}{0,0.28,0.67}

\usepackage[
  breaklinks,
  colorlinks,
  urlcolor=LinkBlue,
  linkcolor=LinkBlue,
  citecolor=LinkBlue
]{hyperref}
\usepackage[scaled]{beramono}
\usepackage[capitalize,noabbrev]{cleveref}

\newcommand{\method}{\textup{\textsc{RigidBench}}\xspace}

\newcommand{\papertitle}{%
  \method: Evaluating Rigid-Body Physics in Video Generation Models%
}

\title{\Large
  \vspace{-2ex}
  \papertitle
}

\author{%
  \small
  Swarnim Jain\thanks{Correspondence to: \href{mailto:swarnimjain004@gmail.com}{\texttt{swarnimjain004@gmail.com}}.},\quad Shangzhe Wu \\
  \small
  University of Cambridge
  \vspace{-0.5ex}
}
\date{}

\hypersetup{
  pdftitle={RigidBench: Evaluating Rigid-Body Physics in Video Generation Models},
  pdfauthor={Swarnim Jain and Shangzhe Wu}
}

\begin{document}

\maketitle

\begin{abstract}
\noindent
Video models are increasingly used to predict what happens next in a scene, yet the metrics commonly used to compare their outputs say little about whether the predicted objects move correctly. Motion, geometry, identity, background stability, and visual similarity can fail independently, but whole-frame scores often mix these errors together. We introduce \method, a simulator-grounded benchmark that compares a generated continuation with a reference rollout from the same initial frame and motion description. Its five rigid-body tasks vary objects, materials, viewpoints, and indoor and outdoor scenes, with per-frame masks, depth, 6-DoF trajectories, and contacts available for scoring. We evaluate eight models on the same 100 examples with ten measurements that keep these aspects separate. The resulting rankings depend strongly on what is measured: no model leads on all ten, and across model means, higher SSIM accompanies larger 3D trajectory error ($r=0.89$). \method also includes 5,000 training videos with exact simulator state, which we use to fine-tune and analyze Wan~2.2 TI2V-5B. Full fine-tuning reduces 3D trajectory error by about 20\% with almost no change in SSIM, while teacher-forced probes and targeted interventions show that object position is represented throughout Wan's diffusion transformer and used by its denoising computation.
\end{abstract}

\section{Introduction}
\label{sec:introduction}

Video models increasingly supply rollouts for robot control, synthetic demonstrations, policy evaluation, and rare driving scenarios~\citep{li2026vera,wang2026interactive,sharma2026worldgymnast,wayve2025gaia3}. When generated frames guide an action or become a training example, the motion they depict matters independently of their visual quality, since a missed surface or an incorrect collision can change the downstream result even when the video looks clean.

Evaluating these rollouts is difficult because a generated video can depart from its reference in several ways at once. An object may follow the wrong path, change shape or identity, or move against a drifting background, and a whole-frame score may respond to all of these changes without identifying any one of them. Existing benchmarks approach the problem through human and vision-language-model judgments~\citep{bansal2025videophy2,li2025worldmodelbench,lin2026phyground}, recorded experiments and conservation laws~\citep{motamed2025physicsiq,tragoudaras2026morpheus}, or controlled simulation~\citep{li2025pisa}. These evaluations reveal broad physical failures, but most report a clip-level judgment or image-space agreement rather than the 3D motion of each object.

\method\footnote{\url{https://github.com/swarnim-j/RigidBench}} controls the comparison by generating a reference rollout in simulation, then giving the video model only its first frame and a short motion description. Since both continuations begin from the same rendered state, the generated video can be compared with aligned masks, depth, 6-DoF trajectories, and contacts rather than with an uninstrumented recording. The five tasks cover free fall, bouncing, a ramp collision, a ball chain, and a drop into a cluster, rendered with varied objects and photorealistic scenes. Together they provide 5,000 training examples and a fixed 100-example evaluation set with held-out scenes and objects.

For each generated video, the evaluator recovers actor masks, point tracks, and depth before applying seven established measurements of mask overlap and shape, image-space motion, depth, and full-frame appearance. We add ATE-3D for world-space trajectories, IdDrift for object appearance along corresponding tracks, and BGDrift for local background deformation after coherent camera motion has been removed. Rather than combine these measurements into one score, we retain them separately so that disagreements between different parts of the prediction remain visible.

Across the eight audited models, visual similarity and trajectory accuracy give markedly different orderings, with higher SSIM accompanying larger 3D trajectory error and no model leading on all ten measurements. Task difficulty also does not follow the number of objects or contacts: free fall produces the largest trajectory error for every model, even though it contains only one moving object, because an early error can persist through most of the rollout.

The same simulator state supports a closer study of Wan~2.2 TI2V-5B. Full fine-tuning lowers ATE-3D by about one fifth while leaving SSIM nearly unchanged, which separates the change in reference motion from the usual measure of frame similarity. Under teacher forcing, linear probes recover position and contact throughout the diffusion transformer, while velocity, orientation, and angular velocity remain weak. Removing the position-aligned directions raises the flow-matching loss far more than matched controls, linking the decoded position signal to Wan's denoising computation.

\section{Related Work}
\label{sec:related-work}

\paragraph{Plausibility judgments.}
Many benchmarks ask whether a generated clip looks physically plausible. VideoPhy-2 tests action-centric physical commonsense, WorldModelBench trains a vision-language model to judge generated videos, and PhyGround combines law-specific questions with controlled human annotation~\citep{bansal2025videophy2,li2025worldmodelbench,lin2026phyground}. This form of evaluation covers many actions and physical phenomena without requiring instrumented scenes, although its scores identify the implausible event or violated rule rather than tracing each object's motion against a reference.

\paragraph{Recorded motion and physical laws.}
Physics-IQ compares generated continuations with repeated recordings of real experiments using motion masks and pixel error~\citep{motamed2025physicsiq}. Morpheus instead estimates physical quantities from controlled experiments and tests generated motion against fitted dynamics and conservation laws, allowing a plausible trajectory to score well without matching one recording exactly~\citep{tragoudaras2026morpheus}. Both benchmarks ground their evaluation in real motion, which also means that the masks, trajectories, and physical quantities needed for scoring must be recovered from video.

\paragraph{Simulator-grounded evaluation.}
Simulation provides exact state and repeatable initial conditions. PhyWorld uses deterministic 2D mechanics to study scaling and generalization, finding that strong in-distribution prediction does not transfer reliably to unseen physical settings~\citep{kang2025physical}, while WorldBench isolates individual concepts and tests whether models recover parameters such as friction and viscosity~\citep{upadhyay2026worldbench}. PISA is closest to our evaluation and training setup, since it measures mask and 2D trajectory agreement for falling objects before post-training a video model on simulated drops~\citep{li2025pisa}. \method extends this approach to object-level 3D motion across five rigid-body tasks, photorealistic scenes, and held-out objects and environments, while reporting appearance and reference motion separately.

\paragraph{Physical information inside video models.}
Recent work uses layerwise probes to recover speed, acceleration, and direction from large video encoders, then applies targeted attention ablations to locate components that support those signals~\citep{joseph2026interpreting}. We study Wan's diffusion transformer using simulator state aligned to each object and time slot, and compare the representations of the base and fine-tuned models. Since a successful probe establishes only that a signal can be decoded, we also remove the position-aligned subspace with Iterative Nullspace Projection and compare the resulting loss change with rank- and variance-matched controls~\citep{ravfogel2020nullspace,elazar2021amnesic}.

\section[RigidBench]{\method}
\label{sec:rigidbench}

\subsection{What We Measure}

Each \method example pairs a controlled simulator rollout with a video-continuation problem: the model receives the first rendered frame and a short description of the motion, then generates the rest of the clip without seeing the remaining simulator frames used for scoring. Ten measurements describe object motion, geometry, identity, background stability, and full-frame similarity, and remain separate so that their disagreements are not hidden inside an aggregate score. Since each input has one simulated continuation, the trajectory measurements report fidelity to that reference rather than physical plausibility in general.

\subsection{Tasks and Ground Truth}

\begin{figure}[H]
  \centering
  \begingroup
  \setlength{\tabcolsep}{2pt}
  \begin{tabular}{@{}ccccc@{}}
    \scriptsize\textbf{Free fall} &
    \scriptsize\textbf{Bounce} &
    \scriptsize\textbf{Ramp collision} &
    \scriptsize\textbf{Drop into cluster} &
    \scriptsize\textbf{Ball chain} \\
    \includegraphics[width=0.185\linewidth]{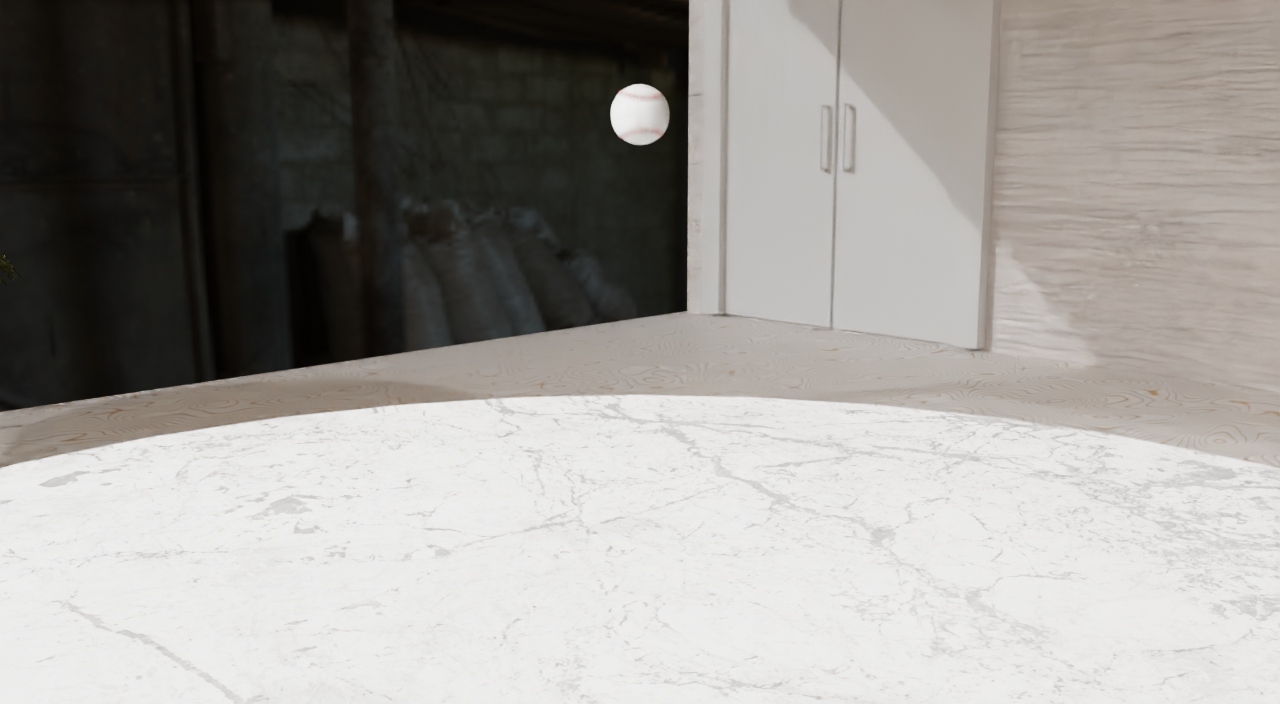} &
    \includegraphics[width=0.185\linewidth]{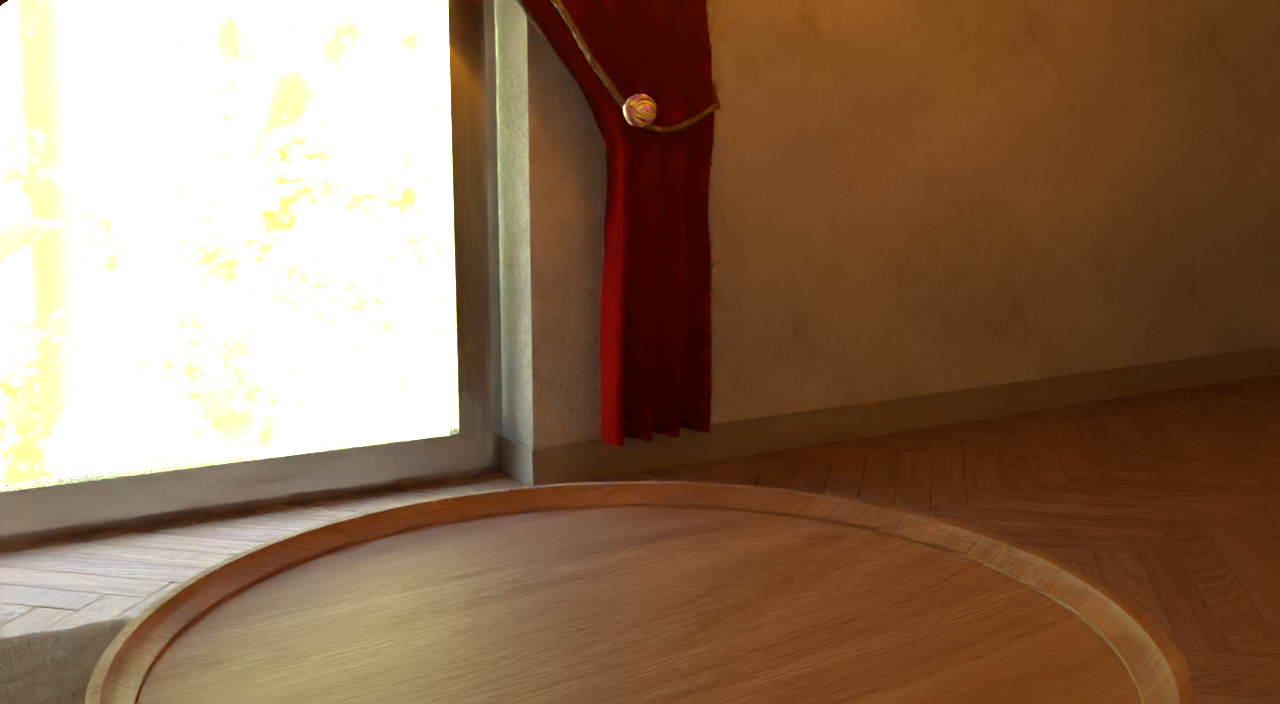} &
    \includegraphics[width=0.185\linewidth]{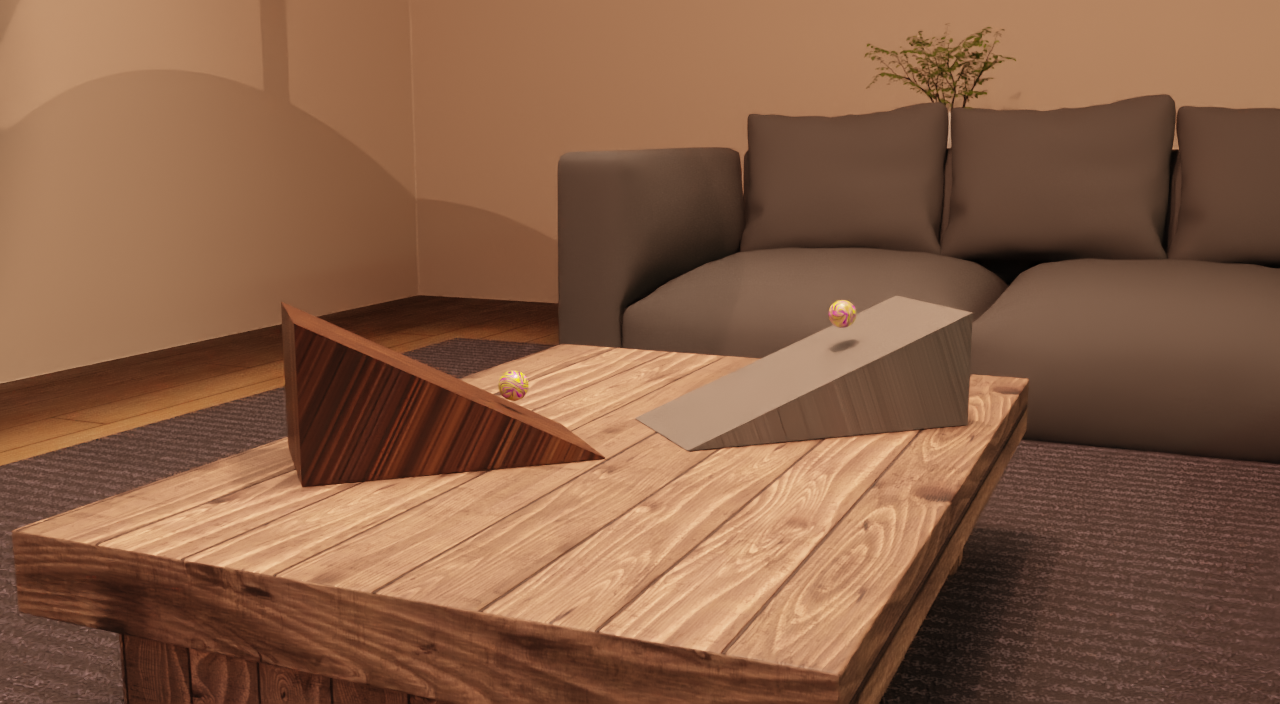} &
    \includegraphics[width=0.185\linewidth]{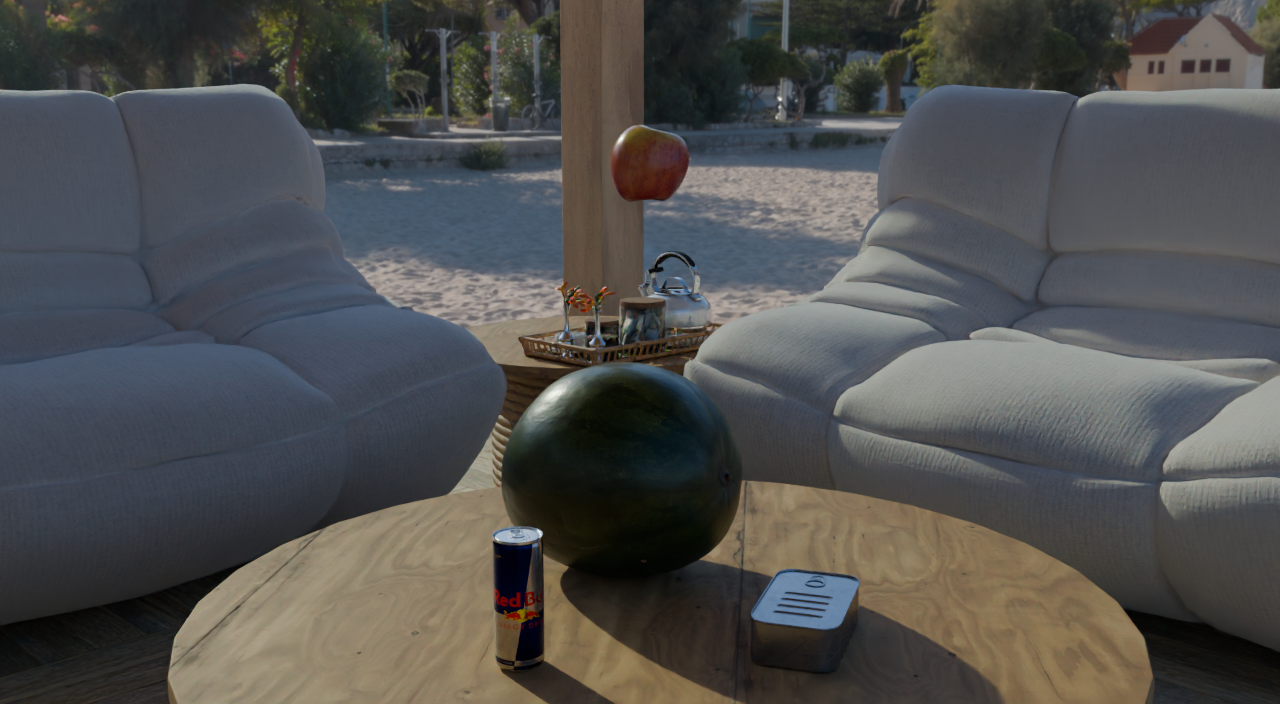} &
    \includegraphics[width=0.185\linewidth]{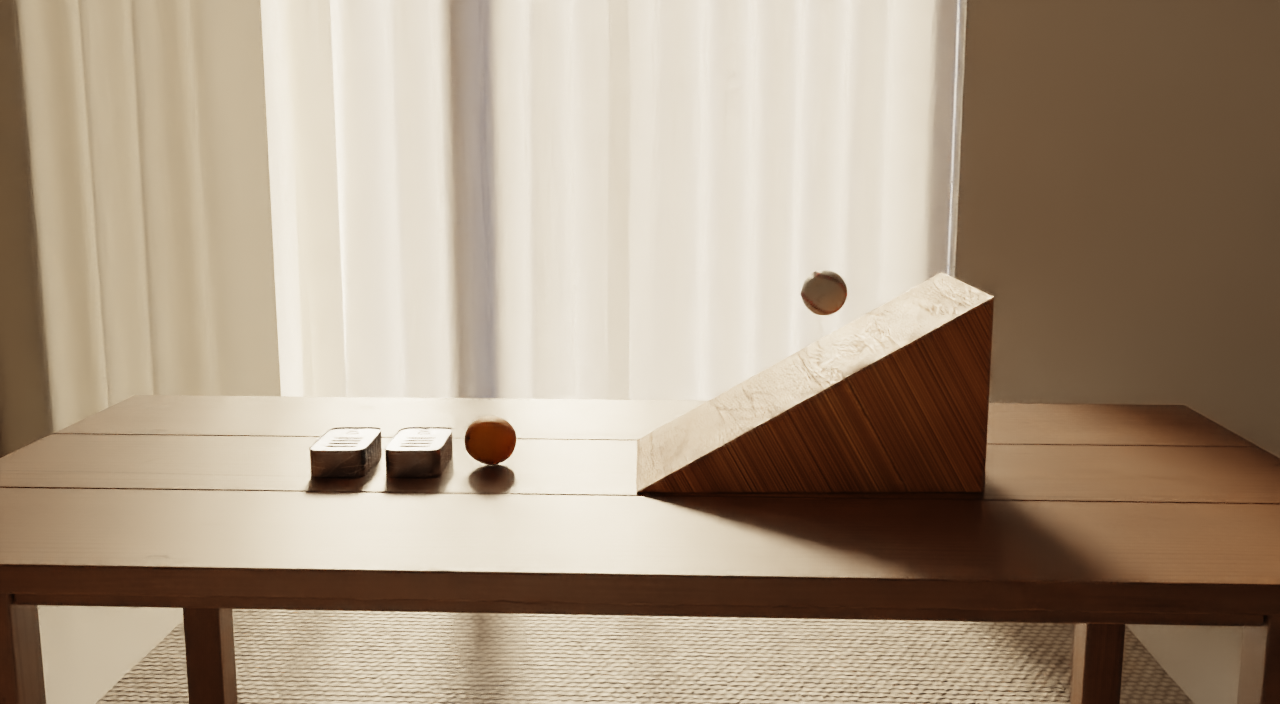}
  \end{tabular}
  \endgroup
  \caption{\textbf{The five \method tasks.} Each image is the first frame of an example in the fixed evaluation set. Objects, scenes, surfaces, dimensions, and materials vary between examples.}
  \label{fig:task-gallery}
\end{figure}

The five tasks cover free flight, restitution, rolling, pairwise collision, and contact cascades, with representative initial frames in \cref{fig:task-gallery} and the corresponding setups in \cref{tab:tasks}. Samples draw from 24 everyday objects, 12 photorealistic indoor and outdoor scenes, and four ramp materials, while varying drop heights, ramp dimensions, material properties, and camera azimuth. These choices vary between examples, but the camera remains fixed within each clip.

\begin{table}[H]
  \centering
  \small
  \setlength{\tabcolsep}{5pt}
  \renewcommand{\arraystretch}{1.08}
  \begin{tabular}{p{0.18\linewidth} p{0.52\linewidth} p{0.21\linewidth}}
    \toprule
    Task & Setup & Main interaction \\
    \midrule
    Free fall & One object drops toward a support & Free flight \\
    Bounce & A high-restitution object drops onto a surface & Restitution \\
    Ramp collision & Two objects roll down opposing ramps and collide & Rolling collision \\
    Ball chain & A rolling object strikes a row of three objects & Contact chain \\
    Drop into cluster & One object falls onto a three-object cluster & Multi-body impact \\
    \bottomrule
  \end{tabular}
  \caption{The five \method tasks. Each is instantiated with different objects, geometry, materials, viewpoints, and scenes.}
  \label{tab:tasks}
\end{table}

Each clip contains 49 frames at $1280\times704$ and 24 frames per second. Blender's Bullet simulator advances the rigid bodies before Cycles renders RGB, a separate mask for each actor, and metric depth from the resulting scene, which keeps all three outputs pixel-aligned. The simulator additionally records each actor's position and orientation, finite-difference linear and angular velocities, and per-frame mesh contacts with their points and normals. A single example seed fixes the sampled assets, physical setup, camera, and render.

\subsection{Training and Evaluation Partitions}

The training manifest contains 5,000 clips across the five tasks, balanced over eight scenes and drawn from 16 objects. The fixed evaluation manifest crosses each task with the four conditions in \cref{tab:splits}, each of which contains five examples per task and therefore 25 examples overall. Four additional scenes and eight additional objects appear only in the evaluation partitions that mark them as held out. Every model is tested on the same frames, prompts, and examples, with the complete task configurations, asset partitions, rendering settings, and ground-truth construction given in \cref{app:benchmark}.

\begin{table}[H]
  \centering
  \small
  \setlength{\tabcolsep}{9pt}
  \begin{tabular}{lccr}
    \toprule
    Condition & Scenes & Objects & Examples \\
    \midrule
    In-distribution & training & training & 25 \\
    OOD-scene & held out & training & 25 \\
    OOD-object & training & held out & 25 \\
    OOD-both & held out & held out & 25 \\
    \bottomrule
  \end{tabular}
  \caption{Evaluation conditions. Every row contains five examples from each task.}
  \label{tab:splits}
\end{table}

\subsection{Metrics and Evaluation}

All models receive the reference first frame, the task description, and the same suffix requesting a locked camera, natural-speed motion, and a continuous take. Before scoring, each output is resized to the reference resolution and aligned to its 24~Hz timestamps, using temporal interpolation for continuous quantities and nearest-neighbor sampling for masks. If a generated clip ends early, scoring stops at its last frame instead of extrapolating it.

Three pretrained models recover the quantities needed for scoring, using the shared first frame to establish the initial actor correspondence. SAM~2 propagates the ground-truth actor masks from that frame~\citep{ravi2025sam2}, CoTracker3 follows 20 points sampled inside each mask~\citep{karaev2025cotracker3}, and Video Depth Anything estimates disparity for every generated frame~\citep{chen2025videodepth}. Seeding masks and tracks in this way avoids adding a separate object-detection and matching problem to the evaluation.

We adapt PISA's mask IoU, height-normalized centroid error, and bidirectional Chamfer distance~\citep{li2025pisa}, and additionally report 2D Absolute Trajectory Error (ATE), scale-invariant depth error (SI-MSE)~\citep{eigen2014depth}, SSIM~\citep{wang2004ssim}, and LPIPS~\citep{zhang2018lpips}. These seven measurements cover mask overlap and shape, image-space motion, depth, and whole-frame appearance, while the remaining three address errors that they do not isolate.

\paragraph{ATE-3D.}
Image-space tracks cannot measure motion along the camera axis on their own, so we fit one scale and shift per clip from predicted disparity to reference disparity before unprojecting each tracked point through the known camera. The geometric median of an actor's visible points gives its reconstructed world position $\widehat{\mathbf p}_{a,t}$. For the valid actor-frame pairs $\mathcal V$, ATE-3D is
\begin{equation}
  \operatorname{ATE\text{-}3D}
  =
  \frac{
    \sqrt{\frac{1}{|\mathcal V|}
      \sum_{(a,t)\in\mathcal V}
      \lVert \widehat{\mathbf p}_{a,t}-\mathbf p_{a,t}\rVert_2^2}
  }{
    \frac{1}{|\mathcal A_m|}
      \sum_{a\in\mathcal A_m}
      \lVert \mathbf p_{a,T-1}-\mathbf p_{a,0}\rVert_2
  },
  \label{eq:ate3d}
\end{equation}
Here $\mathbf p_{a,t}$ is the simulator position and $\mathcal A_m$ contains the actors with non-trivial reference motion, whose mean start-to-end displacement places tasks with different motion ranges on a common scale.

\paragraph{IdDrift.}
For each actor $a$, let $\mathcal P_a$ contain the point-frame pairs that are visible in both clips and yield valid $64\times64$ patches. We center $c^{\mathrm{ref}}_{a,p,t}$ and $c^{\mathrm{gen}}_{a,p,t}$ on the corresponding reference and generated tracks, and embed both with a frozen DINOv2 ViT-L encoder $f$~\citep{oquab2024dinov2}. Then
\begin{equation}
  \begin{aligned}
    s_{a,p,t}
      &= \cos\!\left(f(c^{\mathrm{ref}}_{a,p,t}),
                     f(c^{\mathrm{gen}}_{a,p,t})\right), \\
    \operatorname{IdDrift}
      &= 1 - \frac{1}{|\mathcal A_v|}
        \sum_{a\in\mathcal A_v}
        \frac{1}{|\mathcal P_a|}
        \sum_{(p,t)\in\mathcal P_a} s_{a,p,t},
  \end{aligned}
  \label{eq:iddrift}
\end{equation}
Here $\mathcal A_v$ contains actors with at least one valid patch pair. Each actor receives equal weight so that long or highly visible tracks do not dominate the score, and following corresponding points limits the influence of changes elsewhere in the frame.

\paragraph{BGDrift.}
We track up to 200 corners outside the first-frame actor masks. Let $\mathbf x_i^{(t)}$ be corner $i$ at frame $t$, and let $w_{i,t}$ be the product of its CoTracker3 confidences at frames $0$ and $t$. For each frame, a RANSAC similarity transform $A_t$ maps the frame-zero corners to their current positions. BGDrift is
\begin{equation}
  \operatorname{BGDrift}
  = \frac{1}{H}\,
    \frac{
      \sum_{t=1}^{T-1}\sum_{i\in\mathcal C_t}
      w_{i,t}\left\lVert A_t(\mathbf x_i^{(0)})-\mathbf x_i^{(t)}\right\rVert_2
    }{
      \sum_{t=1}^{T-1}\sum_{i\in\mathcal C_t} w_{i,t}
    },
  \label{eq:bgdrift}
\end{equation}
Here $H$ is the image height and $\mathcal C_t$ contains corners tracked at both endpoints. Since the fitted transform absorbs coherent translation, rotation, and scale, BGDrift measures the local background motion left in the residual.

\section[Video Models on RigidBench]{Video Models on \method}
\label{sec:audit}

Our audit covers eight image-to-video models. We run Wan~2.2 TI2V-5B and Cosmos Predict~2.5-2B locally~\citep{wan2025,cosmos2025}, while Veo~3.1, Veo~3.1 Fast, Kling~3.0, Seedance~2.0, Seedance~2.0 Fast, and Grok Imagine are queried through hosted APIs. Every model is scored on all 100 examples, with the exact identifiers and generation settings recorded in \cref{app:metrics} and further uncertainty analyses reported in \cref{app:audit}.

Since every model sees the same examples, comparisons remain paired until the final average. Model means use 95\% bootstrap intervals from resampled examples, while the cross-model correlation resamples models~\citep{efron1979bootstrap}. Pairwise results report the fraction of matched examples on which one model performs better, following the probability-of-improvement view of evaluation~\citep{agarwal2021rliable}.

\begin{figure}[!t]
  \centering
  \includegraphics[width=\linewidth]{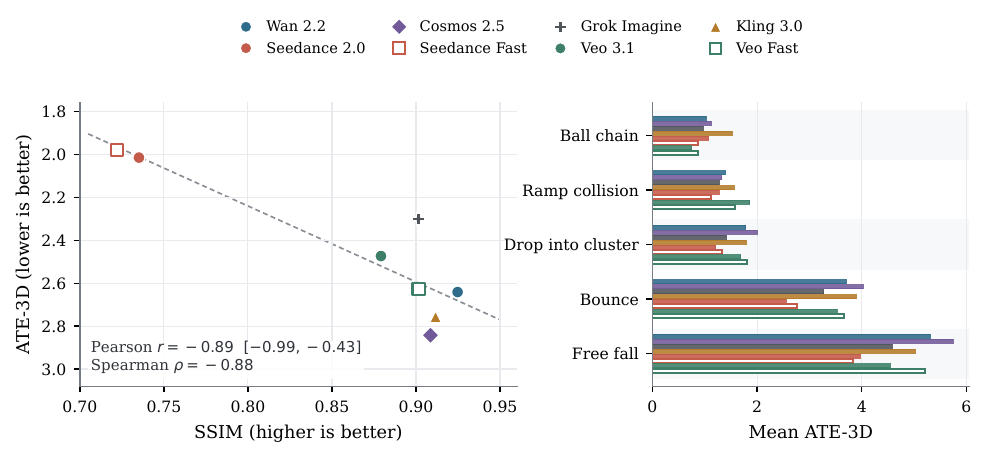}
  \caption{\textbf{Model and task results.} Left: model means on SSIM and ATE-3D. The ATE-3D axis is reversed, and the correlation is computed against negative ATE-3D. Right: mean ATE-3D for each model and task, with one horizontal bar per model. Bars within each task follow the order shown in the key.}
  \label{fig:model-audit}
\end{figure}

\begin{table}[!t]
  \centering
  \footnotesize
  \setlength{\tabcolsep}{3.2pt}
  \renewcommand{\arraystretch}{1.22}
  \resizebox{\linewidth}{!}{%
  \begin{tabular}{lrrrrrrrrrr}
    \toprule
    & \multicolumn{6}{c}{Object motion and geometry} & \multicolumn{4}{c}{Visual similarity and stability} \\
    \cmidrule(lr){2-7}\cmidrule(lr){8-11}
    Model & IoU$\uparrow$ & L2$\downarrow$ & Cham.$\downarrow$ & ATE$\downarrow$ & ATE-3D$\downarrow$ & SI-MSE$\downarrow$ & SSIM$\uparrow$ & LPIPS$\downarrow$ & IdDrift$\downarrow$ & BGDrift$\downarrow$ \\
    \midrule
    Wan~2.2            & .183 & .220 & .370 & .200 & 2.64 & \underline{.0288} & \textbf{.925} & \textbf{.056} & .501 & \underline{.739} \\
    Cosmos~2.5         & .171 & .300 & .527 & .282 & 2.84 & .0331 & .909 & .095 & .484 & 1.094 \\
    Veo~3.1            & .153 & .230 & .380 & .203 & 2.47 & \textbf{.0283} & .879 & .112 & .516 & 2.368 \\
    Veo~3.1 Fast       & .157 & .229 & .382 & .204 & 2.63 & .0352 & .902 & .100 & .517 & 1.491 \\
    Kling~3.0          & .169 & .230 & .387 & \textbf{.186} & 2.76 & .0364 & \underline{.912} & .130 & .553 & 1.632 \\
    Seedance~2.0       & \textbf{.196} & \textbf{.200} & \textbf{.332} & .188 & \underline{2.02} & .0337 & .735 & .145 & \underline{.445} & .845 \\
    Seedance~2.0 Fast  & \underline{.191} & \underline{.203} & \underline{.337} & \underline{.187} & \textbf{1.98} & .0323 & .722 & .157 & \textbf{.444} & \textbf{.579} \\
    Grok Imagine       & .169 & .224 & .373 & .189 & 2.30 & .0362 & .901 & \underline{.081} & .459 & 1.365 \\
    \bottomrule
  \end{tabular}%
  }
  \caption{\textbf{Full model comparison.} Means over the 100-example evaluation set, omitting non-finite measurements. ATE denotes 2D ATE. BGDrift is multiplied by $10^3$. Best and second-best values are bold and underlined.}
  \label{tab:model-results}
\end{table}

\subsection{Model Comparison}

SSIM and ATE-3D produce almost opposite orderings in the left panel of \cref{fig:model-audit}. Across the eight model means, the Pearson correlation between SSIM and negative ATE-3D is $-0.89$, with a bootstrap interval of $[-0.99,-0.43]$, and the Spearman rank correlation is $-0.88$. Wan has the highest SSIM, while the two Seedance variants have both the lowest ATE-3D and the lowest SSIM.

The same reversal holds within matched examples: Seedance~2.0 Fast has lower ATE-3D than every non-Seedance model on at least 65\% of examples, yet both Seedance variants have lower SSIM than every other model on at least 91\%. Because this pattern appears across most individual comparisons, it is not driven by a few unusual clips. Within this audit, full-frame similarity does not identify the models whose objects follow the reference most closely.

The other measurements do not restore a single ordering (\cref{tab:model-results}). Wan leads SSIM and LPIPS, Seedance~2.0 Fast leads ATE-3D, IdDrift, and BGDrift, and Veo~3.1 has the lowest scale-invariant depth error. Since no model leads on all ten, an overall average would hide the tradeoffs shown by the table.

\subsection{Task Breakdown}

Free fall has the largest mean ATE-3D for every model, followed by bounce, as shown in the right panel of \cref{fig:model-audit}. The ordering cannot be explained by the number of objects or contacts, because free fall contains one moving object while the collision tasks contain several. Its distinguishing feature is sustained motion: an early error during an uninterrupted fall changes many later positions, whereas errors in the collision tasks are concentrated around shorter events.

\subsection{Metric Validation}

Controlled perturbations test whether the three new metrics respond to their intended errors. ATE-3D rises equally for displacements along and perpendicular to the camera ray even though 2D ATE misses motion along that ray, while BGDrift removes a coherent image translation but retains independent point jitter with the same mean displacement. IdDrift is $6.3\times$ more sensitive to an identity swap than to a gamma change, compared with $1.3\times$ for pixel L2. The complete curves appear in \cref{app:metrics}.

\section{Fine-tuning and Probing Wan 2.2}
\label{sec:adaptation}

\subsection{LoRA and Full Fine-tuning}

The separate training split lets us ask whether simulator supervision can change the generated motion without substantially changing the video's appearance. We fine-tune Wan~2.2 TI2V-5B on the 5,000 training videos, which are disjoint from the 100 evaluation examples, comparing a rank-32 LoRA run with full DiT fine-tuning. Every saved checkpoint is scored, with the optimization details given in \cref{app:adaptation}.

\begin{figure}[H]
  \centering
  \includegraphics[width=0.86\linewidth]{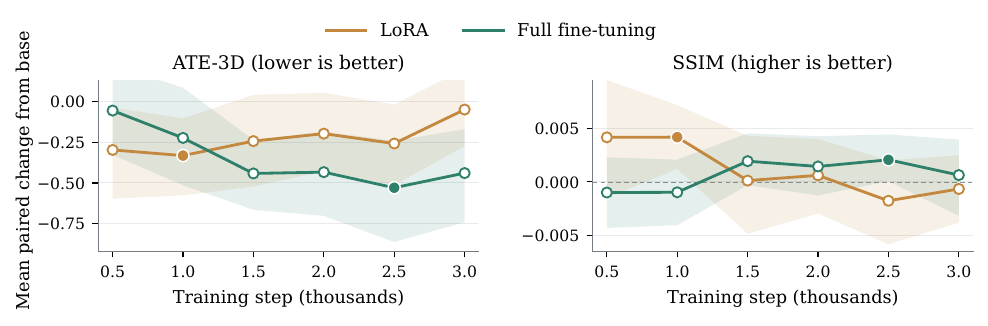}
  \caption{\textbf{Fine-tuning across checkpoints.} Mean paired change from base Wan with bootstrap 95\% confidence intervals. Filled points mark the checkpoint with the lowest ATE-3D in each sweep on this evaluation set.}
  \label{fig:wan-finetuning}
\end{figure}

LoRA improves early before losing much of that gain as training continues, whereas full fine-tuning changes more gradually and reaches an ATE-3D about one fifth below base Wan. Both selected checkpoints improve on the base model, although these examples do not establish an ordering between the two approaches, and the improvement also appears in the image-space trajectory and mask measurements. Appearance scores change little, IdDrift rises slightly under full fine-tuning, and free fall improves most, with the complete sweeps and metric breakdowns reported in \cref{app:adaptation}.

\subsection{Teacher-forced Probes of Physical State}
\label{sec:representations}

The simulator records physical state for every actor and frame, which lets us attach a precise target to Wan's internal features. Following earlier work on diffusion features~\citep{tang2023dift}, we pass noised reference latents through Wan at five noise levels and record the residual stream after every DiT block. Simulator masks identify the tokens belonging to one actor and latent time slot, whose average becomes the hidden vector for that actor at that time. We repeat this extraction for base Wan, the selected full fine-tuning checkpoint, and a random initialization of the same DiT, fitting linear and MLP probes separately at every layer and noise level with splits made by video. Following the logic of control tasks~\citep{hewitt2019control}, we repeat each fit after shuffling its training labels.

\begin{figure}[H]
  \centering
  \includegraphics[width=0.82\linewidth]{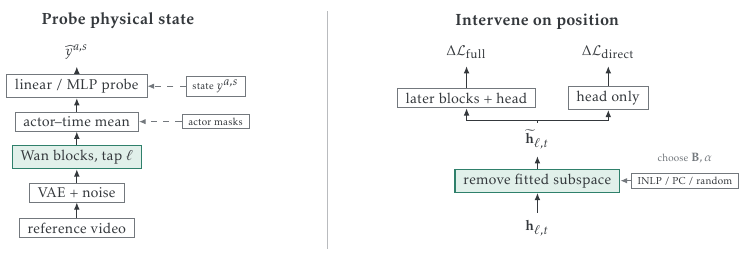}
  \caption{\textbf{Probing and intervention.} Left: actor-aligned hidden states are probed for simulator state. Right: a fitted subspace is removed and Wan's flow-matching loss is measured through two readouts.}
  \label{fig:wan-probe-pipeline}
\end{figure}

Position and contact can be read linearly from the first block, with both signals becoming stronger after fine-tuning. The randomly initialized DiT still scores above chance because its input latent and token layout already contain spatial structure, and an MLP can recover enough of this structure to narrow the difference between all three backbones. The effect of learned features is therefore clearest under a linear readout.

\begin{figure}[H]
  \centering
  \includegraphics[width=0.86\linewidth]{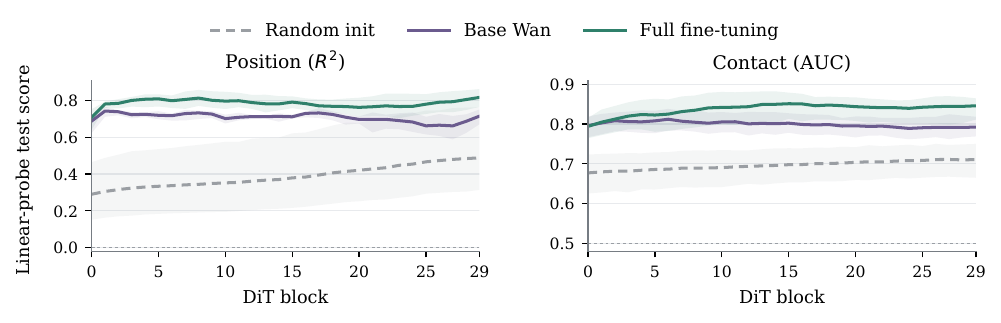}
  \caption{\textbf{Physical state is accessible throughout Wan.} Linear-probe scores by block, averaged over noise levels, with shading spanning their range.}
  \label{fig:wan-probe-scores}
\end{figure}

Among the continuous state variables, only position is recovered strongly across layers and noise levels. Orientation, velocity, angular velocity, and contact normal remain weak, while the acceleration and isolated velocity signals found by nonlinear probes do not persist across the sweep. Wan's residual stream therefore makes current position and contact much easier to recover than instantaneous motion.

A successful probe establishes that a signal can be decoded, but not that Wan uses it in its own computation. We therefore fit an Iterative Nullspace Projection position subspace at each layer and noise level~\citep{ravfogel2020nullspace,elazar2021amnesic}, then intervene as
\begin{equation}
  \widetilde{\mathbf h} = \mathbf h - \alpha \mathbf B\mathbf B^\top\mathbf h,
  \qquad
  \Delta\mathcal L = \mathcal L(\widetilde{\mathbf h})-\mathcal L(\mathbf h).
  \label{eq:amnesic}
\end{equation}
Here $\mathbf B$ contains the INLP directions and $\alpha=1$. The first control removes a random subspace of the same rank, while the second removes principal components scaled to match the centered activation variance removed by the position subspace.

\begin{figure}[H]
  \centering
  \includegraphics[width=0.86\linewidth]{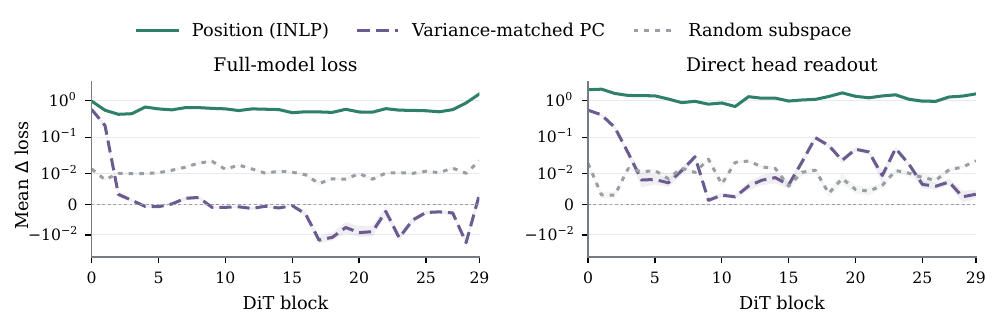}
  \caption{\textbf{Wan depends on position-aligned directions.} Mean loss increase after subspace removal, with bootstrap 95\% confidence intervals. The symmetric-log axis is linear between $-0.01$ and $0.01$.}
  \label{fig:wan-interventions}
\end{figure}

Removing the position-aligned subspace raises the loss far more than either control, both across blocks and after averaging within each sample. This ordering remains unchanged when the modified state passes through the remaining blocks or is read directly by Wan's output head, so neither subspace rank nor removed activation variance accounts for the difference.

Probes of V-JEPA~2 and VideoMAE-v2 recover speed and acceleration early, while motion direction appears around an intermediate-depth transition~\citep{joseph2026interpreting}. Wan has a different profile, with position and contact available from its first block while velocity remains weak, although the experiments also use different units of analysis. That work pools encoder features over a single-object clip, whereas we extract a separate state for each actor and time slot from a generative DiT, then define the removed subspace from the corresponding simulator positions.

\section{Limitations}
\label{sec:limitations}

Each \method input has one controlled reference rollout, although the first frame and prompt do not reveal every state variable or physical parameter. A different continuation may therefore be physically valid without following the same trajectory, so the benchmark measures fidelity to the matched reference rather than every possible future. Its object-level measurements also depend on SAM~2, CoTracker3, Video Depth Anything, and DINOv2. Ground-truth masks and tracks at frame zero establish the initial correspondence, but later segmentation, tracking, or depth errors still limit measurement accuracy.

The current tasks cover rigid bodies in photorealistic synthetic scenes, so real-video evaluation would require separate calibration, while deformable bodies, fluids, and articulated motion would require different state variables and metrics. The audit includes eight related systems, including fast and full variants of the same model families, so its cross-model correlation describes this set rather than video models in general.

\section{Conclusion}
\label{sec:conclusion}

\method shows why video prediction cannot be reduced to one leaderboard: the model that best preserves the frame need not reproduce object motion, and one falling object can accumulate more trajectory error than scenes with several contacts. The Wan~2.2 study carries this distinction inside the model: simulator fine-tuning improves trajectories with little change in SSIM, while the same ground truth identifies a position signal used by denoising, connecting a change in the output to physical information available within the model.

\clearpage
{%
  \small
  \bibliographystyle{plain}
  \bibliography{references}
}

\clearpage
\appendix
\section{Benchmark Construction and Ground Truth}
\label{app:benchmark}

\method samples are generated from declarative files that describe the tasks, scenes, objects, materials, and renderer. This section records the configuration used for the released 5,000-example training set and 100-example evaluation set, whose manifests are fixed before rendering so that a sample identifier determines the task, scene, support surface, allowed objects, and random seed.

\subsection{Tasks}

Each task defines a sequence of placements, a prompt template, the allowed camera directions, and any change to the support surface. Numeric values are sampled uniformly from the stated intervals, while object tags are resolved against the set named by the manifest row. \Cref{tab:task-configs} summarizes the five definitions, and \cref{fig:task-gallery} shows representative evaluation frames.

\begin{table}[H]
  \centering
  \small
  \renewcommand{\arraystretch}{1.16}
  \begin{tabularx}{\linewidth}{@{}p{0.16\linewidth}>{\raggedright\arraybackslash}X>{\raggedright\arraybackslash}p{0.30\linewidth}@{}}
    \toprule
    Task & Physical setup & Prompt and camera direction \\
    \midrule
    Free fall & One object is placed $0.50$--$0.70$\,m above the support. & ``\{object\} is dropping.'' Any azimuth. \\
    Bounce & One object tagged \texttt{bouncy} is placed $0.50$--$0.70$\,m above the support. The support restitution is $0.6$ and its friction is $0.3$. & ``\{object\} is dropping.'' Any azimuth. \\
    Ramp collision & Two opposing ramps each have length $0.35$--$0.45$\,m, height $0.15$--$0.25$\,m, and width $0.20$--$0.30$\,m. Their gap is $0.10$--$0.15$\,m, with one rolling object placed on each ramp. & ``\{object\} and \{object\} are rolling down two ramps facing each other.'' Relative azimuth $70$--$110^\circ$ or $250$--$290^\circ$. \\
    Drop into cluster & Three stackable objects form a cluster with gap $0.003$--$0.008$\,m. A fourth object is placed $0.50$--$0.70$\,m above the support. & ``\{object\} is dropping onto a cluster of objects.'' Any azimuth. \\
    Ball chain & A ramp has length $0.50$--$0.70$\,m, height $0.30$--$0.45$\,m, and width $0.30$--$0.40$\,m. One object rolls toward a line of three objects, placed $0.08$--$0.12$\,m beyond the ramp with gaps of $0.02$--$0.04$\,m. & ``\{object\} is rolling down the ramp toward a row of objects.'' Relative azimuth $60$--$150^\circ$ or $210$--$300^\circ$. \\
    \bottomrule
  \end{tabularx}
  \caption{\textbf{Task configurations.} Camera directions are measured relative to the resolved task rotation. Ramp material is sampled from light wood, dark wood, polished white marble, and rough gray concrete. Prompt placeholders expand to the selected object's name with an indefinite article.}
  \label{tab:task-configs}
\end{table}

The task builder uses four placement operations. It can place an object on a support or at a sampled height above it, create a passive five-face ramp, arrange objects in a line using their bounding boxes, or form a cluster. A cluster places $n$ objects on a ring of radius
\begin{equation}
  R = \frac{r + g/2}{\sin(\pi/n)},
  \label{eq:cluster-radius}
\end{equation}
where $r$ is the largest selected object radius and $g$ is the sampled gap. After placement, a two-dimensional Pymunk simulation runs for 80 steps at $1/60$\,s with zero gravity. Circular footprints repel one another when they overlap, while damped springs keep them near their sampled positions. This short adjustment removes initial intersections before Blender attaches the rigid bodies.

\subsection{Training and Evaluation Partitions}

The asset registry contains 12 scenes, 24 objects, and four ramp materials. The training set uses eight scenes and 16 objects, with the remaining four scenes and eight objects reserved for the held-out evaluation conditions. \Cref{tab:asset-splits} lists the exact identifiers, while each object file specifies its BlenderKit asset, scale, mass, friction, restitution, collision shape, and task tags.

\begin{table}[H]
  \centering
  \small
  \renewcommand{\arraystretch}{1.12}
  \begin{tabularx}{\linewidth}{@{}p{0.19\linewidth}>{\raggedright\arraybackslash}X@{}}
    \toprule
    Subset & Configuration identifiers \\
    \midrule
    Training scenes & \texttt{hallway\_interior}, \texttt{modern\_loft}, \texttt{modern\_luxury\_kitchen}, \texttt{minimal\_white\_kitchen}, \texttt{modern\_japandi\_dining}, \texttt{dining\_room\_set}, \texttt{modular\_flat}, \texttt{coastal\_living\_room} \\
    Held-out scenes & \texttt{beach\_gazebo}, \texttt{butterfly\_mural\_room}, \texttt{cozy\_living\_room}, \texttt{modern\_warm\_kitchen} \\
    Training objects & \texttt{apple}, \texttt{orange}, \texttt{lemon}, \texttt{kiwi}, \texttt{pomegranate}, \texttt{pear}, \texttt{mango}, \texttt{banana}, \texttt{watermelon}, \texttt{baseball}, \texttt{bouncy\_ball}, \texttt{rugby\_ball}, \texttt{red\_bull\_can}, \texttt{plastic\_bottle}, \texttt{large\_dice}, \texttt{food\_can} \\
    Held-out objects & \texttt{peach}, \texttt{tomato}, \texttt{coconut}, \texttt{tennis\_ball}, \texttt{toy\_duck}, \texttt{cricket\_ball}, \texttt{rubiks\_cube}, \texttt{soda\_can} \\
    \bottomrule
  \end{tabularx}
  \caption{\textbf{Scene and object partitions.} The identifiers correspond directly to the released YAML files.}
  \label{tab:asset-splits}
\end{table}

The fixed training manifest contains 5,000 examples, with 625 in each training scene. Its task counts are 1,047 free-fall, 1,000 bounce, 953 ball-chain, 1,000 drop-into-cluster, and 1,000 ramp-collision examples. The evaluation manifest crosses the training and held-out scene and object sets, as shown in \cref{tab:data-partitions}. Each of its four partitions contains five examples from every task, giving 20 examples per task and 100 in total.

\begin{table}[H]
  \centering
  \small
  \renewcommand{\arraystretch}{1.12}
  \begin{tabular}{@{}lccc@{}}
    \toprule
    Partition & Scene set & Object set & Examples \\
    \midrule
    Training & Training & Training & 5,000 \\
    In-distribution & Training & Training & 25 \\
    OOD scene & Held-out & Training & 25 \\
    OOD object & Training & Held-out & 25 \\
    OOD both & Held-out & Held-out & 25 \\
    \bottomrule
  \end{tabular}
  \caption{\textbf{Data partitions.} Each evaluation partition contains five examples from each task. Training examples and evaluation examples are disjoint.}
  \label{tab:data-partitions}
\end{table}

Each manifest row stores the sample identifier, task, scene, support surface, seed, partition, and allowed object identifiers. The seed initializes Python, NumPy, and Cycles, fixing the sampled dimensions, assets, materials, layout adjustment, task rotation, and camera selection.

\subsection{Scene Layout and Camera Selection}

Each scene configuration names one or more planar support surfaces by center, size, and long-axis orientation. Once the task has been assembled, its axis-aligned footprint $(L_T,W_T)$ must fit a support of size $(L_S,W_S)$. For a rotation $\psi$, the condition is
\begin{align}
  L_T\lvert\cos\psi\rvert + W_T\lvert\sin\psi\rvert &\leq L_S, \\
  L_T\lvert\sin\psi\rvert + W_T\lvert\cos\psi\rvert &\leq W_S.
  \label{eq:footprint-fit}
\end{align}
The implementation solves these inequalities over the first quadrant, tiles the valid intervals using the rectangle's symmetries, rotates them into the support's coordinate frame, and samples an angle in proportion to interval length.

A layout that fits the support may still be hidden by furniture or other scene geometry. The camera search therefore tests azimuths every $15^\circ$ within the range allowed by the task. It samples approximately 16 surface points from each target mesh and casts a ray from the candidate camera to every point. The first task rotation with at least 50\% visibility is accepted. If none of eight rotations reaches that threshold, the search uses the pair with the highest visible fraction.

The camera looks at the padded joint bounding box of the actors and support. It uses a $35$\,mm lens, $10^\circ$ elevation, and a distance equal to $1.4$ times the largest framed extent, with a minimum of $1.4$\,m. Depth of field is disabled.

\subsection{Simulation, Rendering, and Stored State}

\Cref{tab:render-config} gives the settings shared by every sample. Blender's Bullet solver advances the rigid bodies before a single Cycles render produces RGB, object index, and depth from the same camera, with the latter two passes written as 32-bit EXR files before conversion to compressed arrays.

\begin{table}[H]
  \centering
  \small
  \renewcommand{\arraystretch}{1.12}
  \begin{tabularx}{\linewidth}{@{}p{0.20\linewidth}>{\raggedright\arraybackslash}X@{}}
    \toprule
    Component & Setting \\
    \midrule
    Video & $1280\times704$, 49 frames, 24 frames per second, PNG frames and H.264 video \\
    Simulation & Blender 4.5.9, 30 Bullet substeps per frame, 20 solver iterations \\
    Rigid bodies & Per-object mass, friction, and restitution, with convex-hull active collisions, linear damping $0.1$, angular damping $0.5$, and collision margin $0.001$\,m \\
    Passive geometry & Mesh collisions with margin $0.0001$\,m, default friction $0.4$, and restitution $0.3$, unless a task overrides the support \\
    Cycles & 32 adaptive samples with threshold $0.05$, OptiX denoising, six maximum bounces, and at most three diffuse, glossy, or transmission bounces \\
    Camera & $35$\,mm lens, $10^\circ$ elevation, $1.4$\,m minimum distance, framing multiplier $1.4$, bounding-box padding $0.2$\,m \\
    Image formation & Motion blur and depth of field disabled, with persistent render data enabled \\
    \bottomrule
  \end{tabularx}
  \caption{\textbf{Simulation and rendering configuration.} These values come from the released render configuration, container image, and Blender setup code.}
  \label{tab:render-config}
\end{table}

One rendered sample contains the files listed in \cref{tab:sample-artifacts}. Its masks, depth, poses, contacts, and RGB frames all come from the same simulated scene.

\begin{table}[H]
  \centering
  \small
  \renewcommand{\arraystretch}{1.12}
  \begin{tabularx}{\linewidth}{@{}p{0.22\linewidth}>{\raggedright\arraybackslash}X@{}}
    \toprule
    Artifact & Contents \\
    \midrule
    \texttt{frames/}, \texttt{video.mp4} & The 49 rendered RGB frames and their encoded video \\
    EXR pass streams & Per-frame 32-bit object-index and Z passes under \texttt{masks/} and \texttt{depth/}, written by the Cycles compositor \\
    \texttt{masks.npz} & Boolean array of shape $(T,N,H,W)$ obtained by thresholding the object-index render pass, together with actor names \\
    \texttt{depth.npz} & Float32 array of shape $(T,H,W)$ containing the Cycles Z pass used by the trajectory code \\
    \texttt{trajectories.npz} & Per-frame world positions, Blender-order quaternions, and finite differences of position and rotation for every actor \\
    \texttt{contacts.json} & Per-frame mesh-overlap events for actor pairs and passive geometry, with up to five contact points and normals per pair \\
    \texttt{metadata.json} & Task, scene, surface, seed, partition, task and camera angles, footprint, actors and roles, camera intrinsics and extrinsics, resolution, frame rate, and prompt \\
    \texttt{prompt.txt} & The instantiated task prompt \\
    \bottomrule
  \end{tabularx}
  \caption{\textbf{Per-sample artifact bundle.} Here $T=49$, $H=704$, $W=1280$, and $N$ is the number of task actors, including passive task geometry such as ramps.}
  \label{tab:sample-artifacts}
\end{table}

\subsection{Ground-truth Point Trajectories}

Image-space trajectories are derived from the frame-zero depth pass and the simulated pose of each rigid body. For a query pixel $\bar{\mathbf u}=[u,v,1]^\top$ with depth $d_0(u,v)$, let $K$ be the camera intrinsic matrix and $S=\operatorname{diag}(1,-1,-1)$ convert computer-vision camera coordinates to Blender camera coordinates. The camera pose $(R_c,\mathbf t_c)$ and the actor pose $(R_{a,0},\mathbf t_{a,0})$ give the body-fixed point
\begin{equation}
  \mathbf p^b = R_{a,0}^{\top}
  \left(R_c S\, d_0(u,v)K^{-1}\bar{\mathbf u}
  + \mathbf t_c - \mathbf t_{a,0}\right).
  \label{eq:body-point}
\end{equation}
At frame $t$, the simulator pose carries this point back into the world, after which it is projected through the fixed camera:
\begin{equation}
  \mathbf p_t^w = R_{a,t}\mathbf p^b + \mathbf t_{a,t},
  \qquad
  \mathbf u_t = \pi\!\left(K S R_c^{\top}(\mathbf p_t^w-\mathbf t_c)\right),
  \label{eq:gt-track}
\end{equation}
where $\pi([X,Y,Z]^\top)=[X/Z,Y/Z]^\top$. Twenty query points are sampled without replacement from a three-pixel erosion of each actor's frame-zero mask. A point remains visible only if it projects in front of the camera, lies inside the actor mask, and its camera depth is no more than $1.01$ times the rendered depth at that pixel. This removes points that later pass behind another object or the task geometry.

\section{Evaluation Details and Metric Validation}
\label{app:metrics}

Every model is evaluated on the same fixed manifest of 100 examples, using the rendered first frame and instantiated task prompt for generation. The same segmentation, tracking, and depth models process every output before the ten metrics are computed, and object-level averages include only actors marked as active in the metadata, excluding ramps and support surfaces.

\subsection{Model Generation and Temporal Alignment}

\Cref{tab:model-generation-settings} records the exact model identifiers and generation settings used for the audit. The two local models run for 50 inference steps in bfloat16, while hosted models are called through Replicate with the first frame resized to $1280\times720$. OpenCV decodes the returned videos, which are then resized with Lanczos interpolation to $1280\times704$.

\begin{table}[H]
  \centering
  \small
  \renewcommand{\arraystretch}{1.12}
  \begin{tabularx}{\linewidth}{@{}p{0.19\linewidth}>{\raggedright\arraybackslash}p{0.34\linewidth}>{\raggedright\arraybackslash}X@{}}
    \toprule
    Model & Identifier & Generation and scoring configuration \\
    \midrule
    Wan 2.2 TI2V-5B & \path{Wan-AI/Wan2.2-TI2V-5B}, local DiffSynth & $1280\times704$, 49 frames at 24\,Hz, 50 steps \\
    Cosmos Predict 2.5-2B & \path{nvidia/Cosmos-Predict2.5-2B}, revision \path{diffusers/base/post-trained}, local Diffusers & $1280\times704$, 33 frames at 16\,Hz, 50 steps \\
    Veo 3.1 / Fast & \path{google/veo-3.1}, \path{google/veo-3.1-fast} & 4\,s, 720p, $16{:}9$, audio disabled, scored as 24\,Hz \\
    Kling 3.0 & \path{kwaivgi/kling-v3-video} & 3\,s, pro mode, $16{:}9$, audio disabled, scored as 24\,Hz \\
    Seedance 2.0 / Fast & \path{bytedance/seedance-2.0}, \path{bytedance/seedance-2.0-fast} & 4\,s, 720p, $16{:}9$, audio disabled, scored as 24\,Hz \\
    Grok Imagine & \path{xai/grok-imagine-video} & 4\,s, 720p, automatic aspect ratio, scored as 24\,Hz \\
    \bottomrule
  \end{tabularx}
  \caption{\textbf{Generation settings.} Model identifiers are shown verbatim. Wan, Cosmos, Veo, and Kling receive a negative prompt. The Seedance and Grok calls do not.}
  \label{tab:model-generation-settings}
\end{table}

We append the following suffix to the task prompt for every model:
\begin{quote}
  \small
  ``Static shot, locked-off camera, fixed tripod, stationary framing. No camera movement, no pan, no tilt, no zoom, no dolly, no drift, no handheld shake. Real-time playback at natural speed, 24 fps, no slow motion, no time-lapse. Realistic rigid-body physics: gravity, momentum, friction, and collisions consistent with reality. Photorealistic, continuous single take, no cuts. Motion begins on the first frame.''
\end{quote}
Models that accept a negative prompt receive:
\begin{quote}
  \small
  ``camera motion, camera shake, pan, tilt, zoom, dolly, parallax, handheld, tracking shot, slow motion, slo-mo, time-lapse, sped-up, motion blur, static start, frozen start, hesitation, delayed action, pause at beginning, floating object, levitation, hovering, teleportation, morphing, warping, deformed geometry, scene cut, jump cut, montage, transition, multiple shots, text, watermark, low quality, low resolution''
\end{quote}

Scoring keeps the reference timestamps $t_k=k/24$ rather than changing the reference video to match each generator. RGB frames, track coordinates, and predicted disparity are linearly interpolated in physical time, while masks and visibility flags use nearest-neighbor sampling. Timestamps beyond the end of a generated video are discarded so that each comparison stops at the shorter sequence. Cosmos is resampled from 16 to 24\,Hz, whereas the hosted clips already run at 24\,Hz and contribute their first 49 frames.

\subsection{Perception and Feature Extraction}

The reference and generated clips share their first frame, so the initial actor correspondence is known and can be used directly by SAM~2 and CoTracker3, as detailed in \cref{tab:perception-settings}. This removes the need for a separate object detector or matching stage.

\begin{table}[H]
  \centering
  \small
  \renewcommand{\arraystretch}{1.12}
  \begin{tabularx}{\linewidth}{@{}p{0.19\linewidth}>{\raggedright\arraybackslash}p{0.30\linewidth}>{\raggedright\arraybackslash}X@{}}
    \toprule
    Quantity & Model & Input and settings \\
    \midrule
    Actor masks & SAM 2.1 Hiera Large, \path{facebook/sam2.1-hiera-large} & One frame-zero mask for each active actor. Propagated logits are thresholded at zero. \\
    Point tracks & CoTracker3 offline & Twenty points per active actor, sampled without replacement from a three-pixel erosion of its reference mask. Generated and reference visibility must both be true. \\
    Disparity & Video Depth Anything Large, \path{depth-anything/Video-Depth-Anything-Large} & ViT-L encoder, input size 384, target rate 24\,Hz. The saved output is disparity. \\
    Identity features & DINOv2 ViT-L/14 & $64\times64$ track-centered patches resized to $224\times224$, ImageNet normalization, bfloat16 inference in batches of 256, and L2-normalized features. \\
    Full-frame features & LPIPS with AlexNet & Reference and generated RGB mapped from $[0,255]$ to $[-1,1]$. \\
    Background tracks & CoTracker3 offline & At most 200 Shi-Tomasi corners with quality threshold $0.01$ and minimum separation 10 pixels, sampled within the background after a 15-pixel erosion. \\
    \bottomrule
  \end{tabularx}
  \caption{\textbf{Perception and feature settings.} Each generated clip is processed independently. The corresponding reference sample supplies the frame-zero inputs.}
  \label{tab:perception-settings}
\end{table}

\subsection{Metric Computation}

\Cref{tab:metric-definitions} summarizes the calculation for one sample, while the equations below give the averaging order and validity conditions. Mask metrics first average active actors within a frame and then average the frames, as do the per-frame values of SSIM, LPIPS, and SI-MSE. ATE instead gives equal weight to every point-frame pair visible in both clips, and non-finite values are omitted from the corresponding mean.

\begin{table}[H]
  \centering
  \small
  \renewcommand{\arraystretch}{1.05}
  \begin{tabularx}{\linewidth}{@{}p{0.18\linewidth}>{\raggedright\arraybackslash}X@{}}
    \toprule
    Metric & Per-sample calculation \\
    \midrule
    IoU $\uparrow$ & Per-actor mask intersection over union, with an empty union assigned one. \\
    Centroid L2 $\downarrow$ & Mask-centroid distance divided by image height. Empty actor-frames are omitted. \\
    Chamfer $\downarrow$ & Sum of the two mean nearest-neighbor distances between foreground-pixel sets, divided by image height. Empty actor-frames are omitted. \\
    ATE $\downarrow$ & Mean track-point distance over jointly visible point-frames, divided by image height. \\
    ATE-3D $\downarrow$ & Actor-position RMSE divided by the mean reference displacement of actors moving more than $0.01$\,m. \\
    SI-MSE $\downarrow$ & Per-frame variance of the log-depth residual after one clip-level affine disparity alignment. \\
    SSIM $\uparrow$ & Mean frame SSIM from \texttt{pytorch\_msssim}, with data range 255. \\
    LPIPS $\downarrow$ & Mean frame LPIPS with an AlexNet backbone. \\
    IdDrift $\downarrow$ & One minus DINOv2 patch cosine similarity, averaged within actors and then across actors, excluding frame zero. \\
    BGDrift $\downarrow$ & Confidence-weighted residual from a per-frame RANSAC similarity transform, divided by image height. \\
    \bottomrule
  \end{tabularx}
  \caption{\textbf{All reported measurements.} Arrows show the direction of better performance. The object-level scores use active actors only.}
  \label{tab:metric-definitions}
\end{table}

\paragraph{Mask and image-space metrics.}
Let $M_{a,t}$ and $\widehat M_{a,t}$ be the reference and generated masks for active actor $a$ at frame $t$, $c(M)$ their pixel centroid, and $S(M)$ their foreground-pixel set. For any per-actor quantity $q_{a,t}$, define the nested mean used by the mask metrics as
\begin{equation}
  \left\langle q_{a,t}\right\rangle_{t,a}
  = \frac{1}{|\mathcal T_q|}
    \sum_{t\in\mathcal T_q}
    \frac{1}{|\mathcal A_{q,t}|}
    \sum_{a\in\mathcal A_{q,t}} q_{a,t},
  \label{eq:appendix-nested-mean}
\end{equation}
where $\mathcal A_{q,t}$ contains actors for which $q_{a,t}$ is defined and $\mathcal T_q$ contains frames with at least one such actor. Also let
\begin{equation}
  d(P,Q)=\frac{1}{|P|}\sum_{p\in P}\min_{q\in Q}\lVert p-q\rVert_2.
\end{equation}
The three mask scores are
\begin{align}
  \operatorname{IoU}
  &= \left\langle
      \frac{|M_{a,t}\cap\widehat M_{a,t}|}
           {|M_{a,t}\cup\widehat M_{a,t}|}
     \right\rangle_{t,a},
  \label{eq:appendix-iou}\\
  \operatorname{L2}
  &= \left\langle
      \frac{\lVert c(M_{a,t})-c(\widehat M_{a,t})\rVert_2}{H}
     \right\rangle_{t,a},
  \label{eq:appendix-l2}\\
  \operatorname{Chamfer}
  &= \left\langle
      \frac{d(S(M_{a,t}),S(\widehat M_{a,t}))
          +d(S(\widehat M_{a,t}),S(M_{a,t}))}{H}
     \right\rangle_{t,a}.
  \label{eq:appendix-chamfer}
\end{align}
An empty IoU union contributes one. Actor-frames for which either mask is empty are omitted from L2 and Chamfer. If $\mathcal Q$ is the set of point-frame pairs visible in both clips, with reference and generated image coordinates $\mathbf u_{p,t}$ and $\widehat{\mathbf u}_{p,t}$, then
\begin{equation}
  \operatorname{ATE}
  = \frac{1}{H|\mathcal Q|}
    \sum_{(p,t)\in\mathcal Q}
    \lVert\widehat{\mathbf u}_{p,t}-\mathbf u_{p,t}\rVert_2.
  \label{eq:appendix-ate}
\end{equation}
Unlike the mask scores, ATE weights every visible point-frame pair equally.

\paragraph{Depth alignment, ATE-3D, and SI-MSE.}
Video Depth Anything predicts disparity up to a clip-dependent affine transformation. Its output is bilinearly resized to the reference grid when necessary. Let $d_t^{\mathrm{pred}}(i,j)$ be predicted disparity and $Z_t^{\mathrm{ref}}(i,j)$ be rendered depth. One scale and offset are fitted over all frames and valid pixels in the clip:
\begin{equation}
  (\widehat{s},\widehat{c})
  = \arg\min_{s,c}
  \sum_{(t,i,j)\in\mathcal V_D}
  \left(s\,d_t^{\mathrm{pred}}(i,j)+c-\frac{1}{Z_t^{\mathrm{ref}}(i,j)}\right)^2,
  \label{eq:appendix-depth-fit}
\end{equation}
where $\mathcal V_D$ contains pixels with finite predicted disparity and reference depth in $(0,100)$\,m. Positive aligned disparities are inverted to obtain $\widehat Z_t$. For the generated points $\mathcal Q_{a,t}$ assigned to actor $a$ and visible at frame $t$, the reconstructed actor position is
\begin{equation}
  \widehat{\mathbf p}_{a,t}
  = \operatorname{gmed}
    \left\{
      \Pi^{-1}\!\left(
        \widehat{\mathbf u}_{p,t},
        \widehat Z_t(\widehat{\mathbf u}_{p,t})
      \right)
      : p\in\mathcal Q_{a,t}
    \right\},
  \label{eq:appendix-reconstruction}
\end{equation}
where $\Pi^{-1}$ uses the known camera intrinsics and extrinsics and depth is sampled bilinearly. The geometric median uses Weiszfeld's algorithm with tolerance $10^{-5}$ and at most 100 iterations. Let $\mathcal V_3$ contain actor-frame pairs with a finite reconstruction, let $\Delta_a=\lVert\mathbf p_{a,T_a-1}-\mathbf p_{a,0}\rVert_2$, and let $\mathcal A_m=\{a:\Delta_a>0.01\text{ m}\}$. The implemented 3D trajectory score is
\begin{equation}
  \operatorname{ATE\text{-}3D}
  = \frac{
      \sqrt{|\mathcal V_3|^{-1}
        \sum_{(a,t)\in\mathcal V_3}
        \lVert\widehat{\mathbf p}_{a,t}-\mathbf p_{a,t}\rVert_2^2}
    }{
      |\mathcal A_m|^{-1}\sum_{a\in\mathcal A_m}\Delta_a
    }.
  \label{eq:appendix-ate3d}
\end{equation}
For SI-MSE, let $\mathcal V_t$ contain pixels with valid reference and aligned predicted depth at frame $t$, and let $\delta_{t,i,j}=\log\widehat Z_t(i,j)-\log Z_t^{\mathrm{ref}}(i,j)$. Then
\begin{equation}
  \operatorname{SI\text{-}MSE}
  = \frac{1}{|\mathcal T_D|}
    \sum_{t\in\mathcal T_D}
    \left[
      \frac{1}{|\mathcal V_t|}\sum_{(i,j)\in\mathcal V_t}\delta_{t,i,j}^2
      -
      \left(
        \frac{1}{|\mathcal V_t|}\sum_{(i,j)\in\mathcal V_t}\delta_{t,i,j}
      \right)^2
    \right],
  \label{eq:appendix-si-mse}
\end{equation}
where $\mathcal T_D$ contains frames with at least one valid pixel.

\paragraph{Full-frame metrics.}
Let $X_t$ and $\widehat X_t$ be the reference and generated RGB frames after temporal and spatial alignment. The reported scores are
\begin{align}
  \operatorname{SSIM}
  &= \frac{1}{T}\sum_{t=0}^{T-1}
     \operatorname{SSIM}_{255}(X_t,\widehat X_t),
  \label{eq:appendix-ssim}\\
  \operatorname{LPIPS}
  &= \frac{1}{T}\sum_{t=0}^{T-1}
     \operatorname{LPIPS}_{\mathrm{AlexNet}}(X_t,\widehat X_t).
  \label{eq:appendix-lpips}
\end{align}
The first operator is the standard single-scale SSIM implementation in \texttt{pytorch\_msssim} with data range 255. LPIPS receives RGB values mapped to $[-1,1]$. Both include frame zero.

\paragraph{IdDrift.}
A patch is retained when at least half of its requested width and height lie inside the image. Reference and generated patches are centered at their corresponding tracks at every jointly visible frame after frame zero. DINOv2 features are normalized before their dot product, and similarities are averaged within an actor before the actor means are averaged. This prevents actors with more visible points from receiving greater weight.
With $\mathcal P_a$ denoting the valid point-frame pairs for actor $a$ and $\mathcal A_v$ the actors with at least one such pair, the exact calculation is
\begin{equation}
  \operatorname{IdDrift}
  = 1-\frac{1}{|\mathcal A_v|}
    \sum_{a\in\mathcal A_v}
    \frac{1}{|\mathcal P_a|}
    \sum_{(p,t)\in\mathcal P_a}
    \cos\!\left(
      f(c^{\mathrm{ref}}_{a,p,t}),
      f(c^{\mathrm{gen}}_{a,p,t})
    \right).
  \label{eq:appendix-iddrift}
\end{equation}

\paragraph{BGDrift.}
The union of the generated frame-zero actor masks defines the foreground. The background is eroded with a $31\times31$ elliptical kernel before corner detection. At every later frame, \texttt{estimateAffinePartial2D} fits translation, rotation, and uniform scale between the frame-zero corners and their current positions. Corner weights are the product of their CoTracker3 confidences at the two frames. At least four corners are required. If RANSAC does not return a transform, raw displacement is used for that frame.
Writing $A_t$ for the fitted transform, $\mathbf x_i^{(t)}$ for a corner position, $w_{i,t}$ for its confidence product, and $\mathcal C_t$ for corners with positive confidence at both endpoints gives
\begin{equation}
  \operatorname{BGDrift}
  = \frac{1}{H}
    \frac{
      \sum_{t=1}^{T-1}\sum_{i\in\mathcal C_t}
      w_{i,t}\lVert A_t(\mathbf x_i^{(0)})-\mathbf x_i^{(t)}\rVert_2
    }{
      \sum_{t=1}^{T-1}\sum_{i\in\mathcal C_t}w_{i,t}
    }.
  \label{eq:appendix-bgdrift}
\end{equation}

\subsection{Aggregation}

Each example produces a JSON file of scalar measurements and an NPZ file containing available per-frame arrays. The released summary reports the arithmetic mean, population standard deviation, and finite count for each metric, followed by the same aggregation within each task. The confidence intervals and paired comparisons applied to these per-example results are reported with the extended model results in \cref{app:audit}.

\subsection{Controlled Metric Perturbations}

We test the three new metrics by changing one known quantity while holding a matched alternative fixed. \Cref{fig:metric-stress-tests} plots the resulting responses.

\providecommand{\metricstressmaxheight}{\textheight}
\begin{figure}[H]
  \centering
  \includegraphics[width=\linewidth,height=\metricstressmaxheight,keepaspectratio]{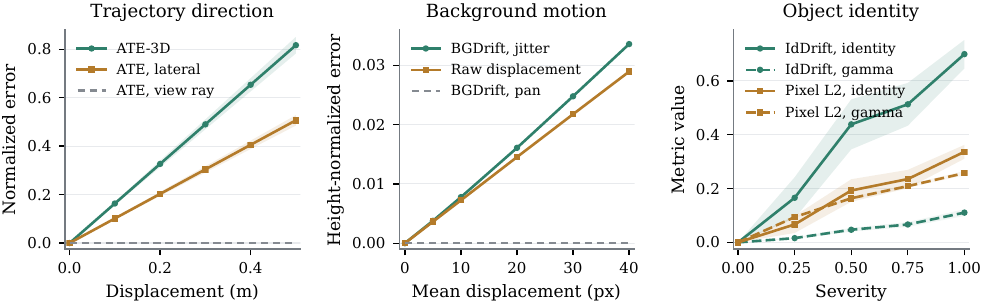}
  \caption{\textbf{Controlled metric perturbations.} Left: equal 3D offsets are applied along the viewing ray and perpendicular to it. Center: coherent translation and independent corner jitter have matched raw displacement. Right: patches undergo an identity swap or a gamma change. Shading gives one standard error from the saved test output.}
  \label{fig:metric-stress-tests}
\end{figure}

For ATE-3D, ten examples, split evenly between free fall and ramp collision, are perturbed at every frame by $0$--$0.5$\,m. Moving the actor along the viewing ray leaves its image coordinate unchanged, so 2D ATE remains zero, whereas ATE-3D responds at the same rate as it does to a perpendicular displacement of equal magnitude.

For BGDrift, 200 corners are followed across 49 frames under coherent translation and independent Gaussian jitter at matched mean displacements from 0 to 40 pixels. The fitted transform removes the translation, while its residual tracks the jitter. For IdDrift, ten frame-zero patches are tested over 20 trials. At severity $q$, the target swaps a patch with probability $q$, while the foil changes the same patch with $\gamma=1+1.5q$. IdDrift's increase at maximum severity is $6.3$ times larger for the identity swap than for the gamma change, compared with $1.3$ times for pixel L2.

\section{Extended Model Results}
\label{app:audit}

\subsection{Uncertainty Across Metrics}

\begin{figure}[H]
  \centering
  \includegraphics[width=\linewidth]{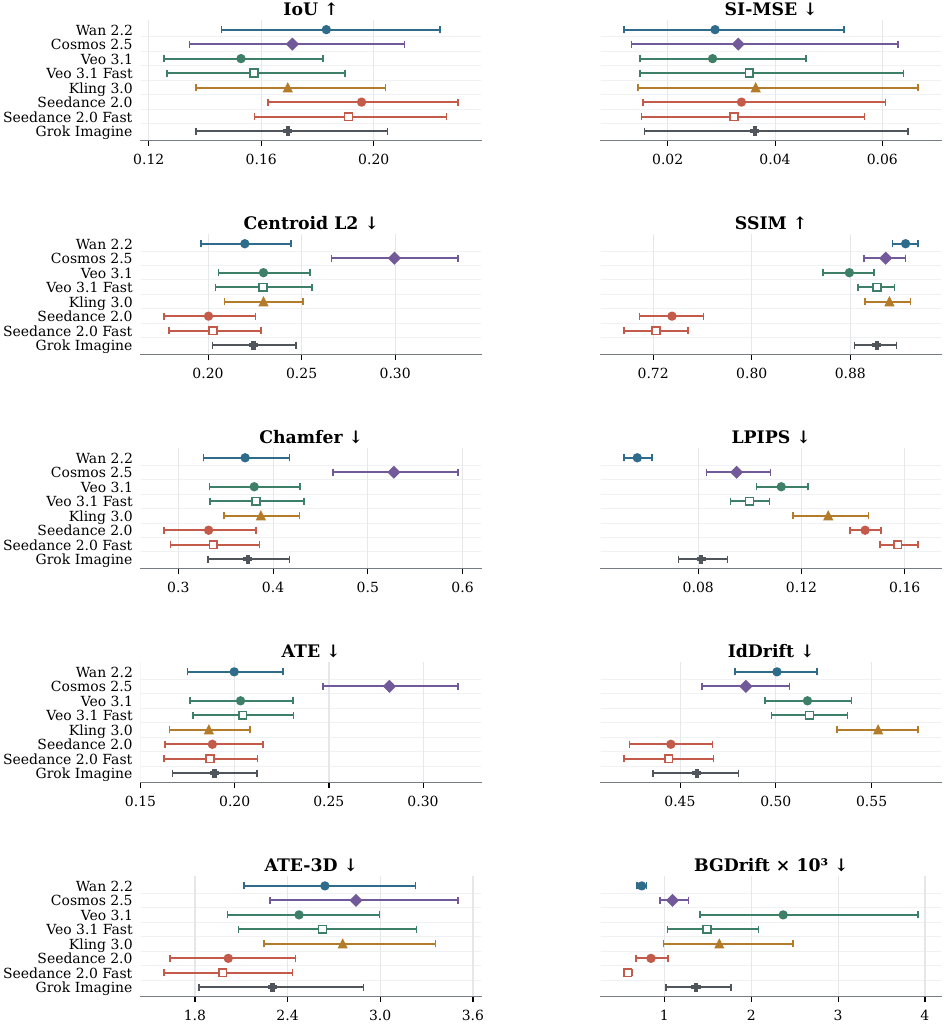}
  \caption{\textbf{Uncertainty across all ten measurements.} Points are model means and bars are bootstrap 95\% confidence intervals over evaluation examples. Models follow the same fixed row order in every panel. BGDrift is multiplied by $10^3$.}
  \label{fig:all-metric-intervals}
\end{figure}

All intervals in this section use 20,000 percentile bootstrap resamples with the example as the resampling unit, preserving the frames, actors, and tracks that contribute to one per-example score. \Cref{fig:all-metric-intervals} applies the same calculation to every model and metric in \cref{tab:model-results}. All cells contain 100 finite values except Veo~3.1 on IdDrift, for which two clips contain no valid patch pairs and are omitted.

Trajectory and depth measurements have wider intervals than the full-frame appearance scores, and some neighboring model means cannot be separated on a single metric. The larger pattern remains clear, however, because the models with the lowest trajectory errors are not those with the highest SSIM or lowest LPIPS.

\subsection{Paired Model Comparisons}

Because every model is evaluated on the same examples, a comparison can be made before averaging. For model $i$ against model $j$, let $w_k$ be one when $i$ is better on example $k$, one half for a tie, and zero otherwise. We report
\begin{equation}
  \operatorname{PoI}(i,j)=\frac{1}{N}\sum_{k=1}^{N} w_k,
  \label{eq:probability-improvement}
\end{equation}
and bootstrap the matched $w_k$ values. This statistic asks how often one model improves on another without allowing a few large errors to determine the comparison.

\begin{figure}[H]
  \centering
  \includegraphics[width=\linewidth]{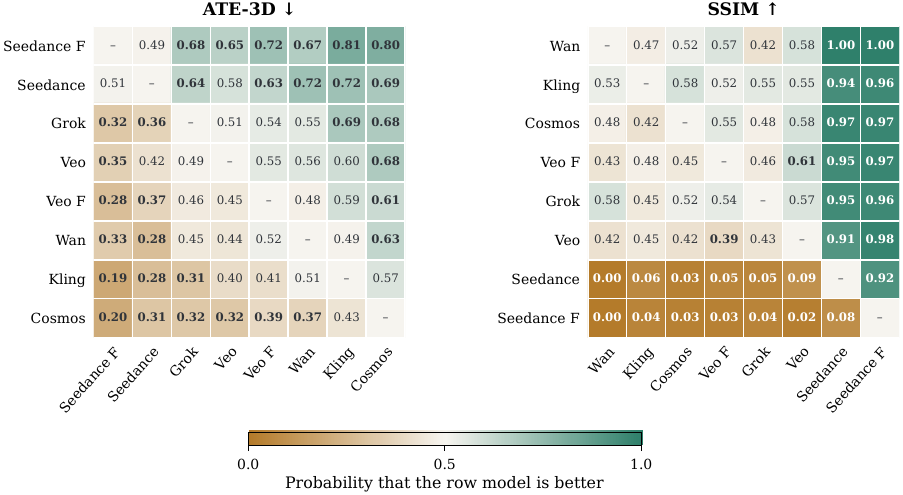}
  \caption{\textbf{Matched probability of improvement.} Each cell is the probability that the row model is better than the column model on the same example. Rows and columns are ordered by the corresponding model mean, best first. Bold entries have a paired bootstrap 95\% confidence interval that excludes $0.5$.}
  \label{fig:pairwise-probabilities}
\end{figure}

The two matrices reverse the models at their extremes. Each Seedance variant beats every non-Seedance model on ATE-3D in at least 65\% of examples. On SSIM, every non-Seedance model beats both variants in at least 91\%. The difference seen in the model means therefore also appears within matched examples.

\subsection{Task Breakdown}

\begin{table}[H]
  \centering
  \small
  \setlength{\tabcolsep}{4.2pt}
  \renewcommand{\arraystretch}{1.12}
  \begin{tabular*}{\linewidth}{@{\extracolsep{\fill}}lrrrrr@{}}
    \toprule
    Model & Ball chain & Ramp collision & Drop into cluster & Bounce & Free fall \\
    \midrule
    Wan~2.2 & 1.02 & 1.39 & 1.78 & 3.70 & 5.30 \\
    Cosmos~2.5 & 1.12 & 1.32 & 2.01 & 4.02 & 5.74 \\
    Veo~3.1 & \textbf{0.75} & 1.86 & 1.69 & 3.53 & 4.54 \\
    Veo~3.1 Fast & \underline{0.86} & 1.59 & 1.80 & 3.66 & 5.22 \\
    Kling~3.0 & 1.52 & 1.56 & 1.79 & 3.89 & 5.02 \\
    Seedance~2.0 & 1.07 & \underline{1.27} & \textbf{1.21} & \textbf{2.55} & \underline{3.97} \\
    Seedance~2.0 Fast & 0.87 & \textbf{1.12} & \underline{1.32} & \underline{2.76} & \textbf{3.83} \\
    Grok Imagine & 0.97 & 1.28 & 1.41 & 3.27 & 4.58 \\
    \bottomrule
  \end{tabular*}
  \caption{\textbf{ATE-3D by task.} Each cell is the mean over the 20 examples for that task. Lower is better; best and second-best values are bold and underlined.}
  \label{tab:task-ate3d}
\end{table}

Free fall has the highest ATE-3D for all eight models, while ball chain has the lowest for all eight. The shared ordering follows the duration of motion more closely than the number of objects or contacts. Position errors accumulate throughout a fall, whereas much of a collision clip occurs before or after a short interaction.

\section{Fine-tuning Wan 2.2}
\label{app:adaptation}

\subsection{Training Data and Objective}

Both runs start from Wan~2.2 TI2V-5B and use the 5,000-example training manifest from \cref{app:benchmark}, which is disjoint from the evaluation manifest. Before optimization, Wan's T5 model encodes the prompts and its VAE encodes the videos. The dataset contains 624 distinct prompts, so repeated prompts reuse the same text embedding, while all 49 frames are resized from $1280\times704$ to $640\times352$ before VAE encoding produces 13 temporal latent slots. The float16 text embeddings and video latents are streamed during training without spatial or temporal augmentation, and evaluation returns to the benchmark resolution of $1280\times704$.

Each update starts from a ground-truth video latent $\mathbf z$ and samples one index $t$ uniformly from the scheduler's 1,000 training timesteps for each local batch. With Gaussian noise $\boldsymbol\epsilon$, the scheduler produces a noised latent $q_t(\mathbf z,\boldsymbol\epsilon)$ and its flow target $\mathbf v_t$. The first temporal slot is then restored to its clean value:
\begin{equation}
  \widetilde{\mathbf z}_t=q_t(\mathbf z,\boldsymbol\epsilon),
  \qquad
  [\widetilde{\mathbf z}_t]_{:,:,0:1}=[\mathbf z]_{:,:,0:1}.
  \label{eq:finetuning-noise}
\end{equation}
The DiT predicts velocity from $\widetilde{\mathbf z}_t$, the timestep, and the cached text context $\mathbf c$. The first slot is also excluded from the float32 MSE:
\begin{equation}
  \mathcal L(\theta)=w(t)\,
  \operatorname{MSE}\!\left(
    [f_\theta(\widetilde{\mathbf z}_t,t,\mathbf c)]_{:,:,1:},
    [\mathbf v_t(\mathbf z,\boldsymbol\epsilon)]_{:,:,1:}
  \right),
  \label{eq:finetuning-loss}
\end{equation}
where $w(t)$ is Wan's scheduler weight. This is the same first-frame pinning used when the resulting checkpoints generate videos for evaluation.

\subsection{Optimization and Parameter Updates}

Full fine-tuning updates every DiT parameter except the text and time conditioning modules, which remain fixed. LoRA freezes the base DiT and inserts rank-32 updates into the attention and feed-forward projections of every block. \Cref{tab:finetuning-config} compares the two configurations.

\begin{table}[H]
  \centering
  \small
  \renewcommand{\arraystretch}{1.13}
  \begin{tabularx}{\linewidth}{@{}p{0.25\linewidth}>{\raggedright\arraybackslash}X>{\raggedright\arraybackslash}X@{}}
    \toprule
    Setting & LoRA & Full fine-tuning \\
    \midrule
    Trainable weights & Rank $32$, $\alpha=32$, dropout $0$ & All non-conditioning DiT weights \\
    Updated modules & \texttt{q}, \texttt{k}, \texttt{v}, \texttt{o}, \texttt{ffn.0}, \texttt{ffn.2} & All remaining DiT modules; \texttt{text\_embedding}, \texttt{time\_embedding}, and \texttt{time\_projection} fixed \\
    Distributed strategy & DDP & FSDP \texttt{FULL\_SHARD}, wrapped by DiT block \\
    Batch per GPU & 2 & 16 \\
    Gradient accumulation & 8 batches & 1 batch \\
    Effective batch & 128 examples & 128 examples \\
    Learning rate & $1\times10^{-4}$ & $2\times10^{-5}$ \\
    Cosine schedule & 5,000-step horizon, 250-step warmup & 3,000-step horizon, 150-step warmup \\
    Gradient checkpointing & Disabled & Enabled \\
    Evaluated checkpoints & Every 500 steps through step 3,000 & Every 500 steps through step 3,000 \\
    \bottomrule
  \end{tabularx}
  \caption{\textbf{Fine-tuning configurations.} Both runs use eight H100 GPUs, bf16 mixed precision, fused AdamW with weight decay $0.01$, $\epsilon=10^{-8}$, default $\beta$ values, gradient clipping at $1.0$, and random seed 42.}
  \label{tab:finetuning-config}
\end{table}

The LoRA run uses DDP because only the adapter tensors need gradients and optimizer state. Full fine-tuning uses FSDP to shard model weights, gradients, and AdamW state across the eight GPUs.

Checkpoints are written every 500 updates. The LoRA configuration has a 5,000-step scheduler horizon, but the saved and evaluated run ends at step 3,000.

\subsection{Complete Checkpoint Results}

\begin{table}[H]
  \centering
  \scriptsize
  \setlength{\tabcolsep}{2.6pt}
  \renewcommand{\arraystretch}{1.12}
  \resizebox{\linewidth}{!}{%
  \begin{tabular}{llrrrrrrrrrr}
    \toprule
    & & \multicolumn{6}{c}{Object motion and geometry} & \multicolumn{4}{c}{Visual similarity and stability} \\
    \cmidrule(lr){3-8}\cmidrule(lr){9-12}
    Regime & Step & IoU$\uparrow$ & L2$\downarrow$ & Cham.$\downarrow$ & ATE$\downarrow$ & ATE-3D$\downarrow$ & SI-MSE$\downarrow$ & SSIM$\uparrow$ & LPIPS$\downarrow$ & IdDrift$\downarrow$ & BGDrift$\downarrow$ \\
    \midrule
    Base & -- & .183 & .220 & .370 & .200 & 2.64 & .0288 & .925 & .056 & \textbf{.501} & .739 \\
    \addlinespace[2pt]
    LoRA & 500 & .185 & .168 & .272 & .158 & 2.34 & \textbf{.0247} & \underline{.929} & \underline{.056} & .525 & .726 \\
    LoRA & \textbf{1,000} & \textbf{.207} & .164 & .264 & .154 & 2.31 & \underline{.0250} & \textbf{.929} & \textbf{.056} & \underline{.518} & \underline{.718} \\
    LoRA & 1,500 & .195 & .174 & .282 & .163 & 2.40 & .0261 & .925 & .058 & .519 & .770 \\
    LoRA & 2,000 & .193 & .165 & .264 & .159 & 2.44 & .0260 & .925 & .058 & .523 & .736 \\
    LoRA & 2,500 & .194 & \underline{.163} & \underline{.261} & .153 & 2.38 & .0277 & .923 & .060 & .527 & .737 \\
    LoRA & 3,000 & .186 & .171 & .276 & .163 & 2.59 & .0274 & .924 & .060 & .522 & .762 \\
    \addlinespace[2pt]
    Full FT & 500 & .185 & .187 & .306 & .159 & 2.59 & .0288 & .924 & .062 & .518 & \textbf{.717} \\
    Full FT & 1,000 & .190 & .171 & .273 & .150 & 2.42 & .0298 & .924 & .064 & .527 & .745 \\
    Full FT & 1,500 & .191 & .168 & .270 & \underline{.147} & \underline{2.20} & .0274 & .927 & .062 & .520 & .733 \\
    Full FT & 2,000 & .193 & \textbf{.160} & \textbf{.253} & \textbf{.137} & 2.21 & .0280 & .926 & .061 & .529 & .756 \\
    Full FT & \textbf{2,500} & \underline{.199} & .165 & .264 & .150 & \textbf{2.11} & .0262 & .927 & .058 & .523 & .737 \\
    Full FT & 3,000 & .193 & .174 & .281 & .150 & 2.20 & .0256 & .925 & .060 & .525 & .755 \\
    \bottomrule
  \end{tabular}%
  }
  \caption{\textbf{Complete checkpoint sweep.} Means over the 100-example evaluation set. Bold checkpoint labels mark the minimum ATE-3D within each run and are the checkpoints used in the subsequent comparisons. Within each metric, the best and second-best values are bold and underlined. BGDrift is multiplied by $10^3$.}
  \label{tab:finetuning-sweep}
\end{table}

The checkpoints marked in \cref{tab:finetuning-sweep} have the lowest ATE-3D in their respective saved sweeps. The same 100 examples are used for this selection and for the reported metrics. These are therefore descriptions of the observed sweeps rather than held-out checkpoint-selection estimates. \Cref{tab:finetuning-paired-changes} gives paired bootstrap intervals from 20,000 resamples of the matched examples.

\begin{table}[H]
  \centering
  \footnotesize
  \setlength{\tabcolsep}{3.5pt}
  \renewcommand{\arraystretch}{1.15}
  \begin{tabular*}{\linewidth}{@{\extracolsep{\fill}}lccc@{}}
    \toprule
    Metric & LoRA 1k $-$ base & Full FT 2.5k $-$ base & Full FT $-$ LoRA \\
    \midrule
    IoU $\uparrow$ & $+.024\;[+.008,+.041]$ & $+.016\;[+.001,+.030]$ & $-.009\;[-.027,+.009]$ \\
    L2 $\downarrow$ & $-.056\;[-.076,-.036]$ & $-.055\;[-.076,-.035]$ & $+.001\;[-.019,+.021]$ \\
    Cham. $\downarrow$ & $-.106\;[-.146,-.068]$ & $-.107\;[-.148,-.067]$ & $-.000\;[-.039,+.038]$ \\
    ATE $\downarrow$ & $-.045\;[-.064,-.027]$ & $-.050\;[-.069,-.031]$ & $-.004\;[-.024,+.015]$ \\
    ATE-3D $\downarrow$ & $-.33\;[-.58,-.10]$ & $-.53\;[-.86,-.25]$ & $-.20\;[-.50,+.07]$ \\
    SI-MSE $\downarrow$ & $-.00376\;[-.00741,-.00091]$ & $-.00262\;[-.00583,-.00002]$ & $+.00114\;[+.00005,+.00243]$ \\
    SSIM $\uparrow$ & $+.0042\;[+.0012,+.0072]$ & $+.0021\;[+.0000,+.0044]$ & $-.0021\;[-.0042,+.0004]$ \\
    LPIPS $\downarrow$ & $-.0006\;[-.0031,+.0016]$ & $+.0020\;[-.0001,+.0040]$ & $+.0026\;[+.0007,+.0048]$ \\
    IdDrift $\downarrow$ & $+.017\;[-.002,+.036]$ & $+.022\;[+.001,+.045]$ & $+.005\;[-.011,+.022]$ \\
    BGDrift $\downarrow$ & $-.021\;[-.050,+.007]$ & $-.001\;[-.038,+.042]$ & $+.020\;[-.007,+.051]$ \\
    \bottomrule
  \end{tabular*}
  \caption{\textbf{Paired changes for the selected checkpoints.} Each entry is the mean paired change followed by its bootstrap 95\% confidence interval. Positive is better for IoU and SSIM; negative is better for every other row. BGDrift is multiplied by $10^3$.}
  \label{tab:finetuning-paired-changes}
\end{table}

Both selected checkpoints improve all six motion and geometry measurements relative to base Wan. SSIM also rises slightly, while the LPIPS and BGDrift intervals include zero for both runs. Full fine-tuning raises IdDrift by $0.022$ with an interval of $[0.001,0.045]$, whereas the LoRA change is not resolved. The ATE-3D difference between the selected full and LoRA checkpoints is $-0.20$ with an interval of $[-0.50,0.07]$, so their ordering is not established by these examples.

\subsection{Task and Evaluation-Partition Breakdowns}

\begin{table}[H]
  \centering
  \small
  \setlength{\tabcolsep}{4.2pt}
  \renewcommand{\arraystretch}{1.12}
  \begin{tabular*}{\linewidth}{@{\extracolsep{\fill}}lrrrrr@{}}
    \toprule
    Model & Ball chain & Ramp collision & Drop into cluster & Bounce & Free fall \\
    \midrule
    Base & 1.02 & 1.39 & 1.78 & 3.70 & 5.30 \\
    LoRA 1k & \underline{0.94} & \underline{1.14} & \textbf{1.41} & \textbf{3.20} & \underline{4.85} \\
    Full FT 2.5k & \textbf{0.78} & \textbf{1.12} & \underline{1.51} & \underline{3.21} & \textbf{3.94} \\
    \bottomrule
  \end{tabular*}
  \caption{\textbf{Selected checkpoints by task.} Mean ATE-3D over 20 examples per task. Lower is better; best and second-best values are bold and underlined.}
  \label{tab:finetuning-tasks}
\end{table}

\begin{table}[H]
  \centering
  \small
  \setlength{\tabcolsep}{3.4pt}
  \renewcommand{\arraystretch}{1.12}
  \begin{tabular*}{\linewidth}{@{\extracolsep{\fill}}lrrrrrrrr@{}}
    \toprule
    & \multicolumn{4}{c}{ATE-3D $\downarrow$} & \multicolumn{4}{c}{SSIM $\uparrow$} \\
    \cmidrule(lr){2-5}\cmidrule(lr){6-9}
    Model & ID & Scene & Object & Both & ID & Scene & Object & Both \\
    \midrule
    Base & 2.64 & 2.31 & 3.63 & 1.98 & 0.946 & 0.906 & 0.946 & 0.901 \\
    LoRA 1k & \underline{2.20} & \underline{2.18} & \underline{3.36} & \underline{1.50} & \textbf{0.948} & \textbf{0.916} & \underline{0.946} & \textbf{0.906} \\
    Full FT 2.5k & \textbf{1.95} & \textbf{1.99} & \textbf{3.03} & \textbf{1.48} & \underline{0.947} & \underline{0.910} & \textbf{0.948} & \underline{0.902} \\
    \bottomrule
  \end{tabular*}
  \caption{\textbf{Selected checkpoints by evaluation partition.} ID uses training scenes and objects; Scene, Object, and Both use the corresponding held-out sets. Each cell averages 25 examples. Best and second-best values are bold and underlined.}
  \label{tab:finetuning-splits}
\end{table}

Both selected checkpoints reduce mean ATE-3D on every task and evaluation partition. Their relative ordering varies by task: full fine-tuning is lower on ball chain, ramp collision, and free fall, while LoRA is slightly lower on drop into cluster and bounce. Free fall accounts for the largest absolute change under full fine-tuning. Neither selected checkpoint lowers the split-mean SSIM relative to base Wan.

\section{Probe and Intervention Details}
\label{app:representations}

\subsection{Actor-Aligned Residual Streams}

The activation sweep uses the 5,000 simulator-generated training videos rather than the 100 evaluation examples, splitting them into 4,000 for probe training, 500 for validation, and 500 for testing. Since the full fine-tuned backbone was trained on the same corpus, its probe scores describe access to state on that training distribution.

Each 49-frame video is resized to $352\times640$ and encoded once with Wan's VAE. This gives 13 temporal latent slots on a $44\times80$ spatial grid, which Wan patchifies into a $13\times22\times40$ residual-stream grid with width 3,072. We add noise at five positions $\tau\in\{0.1,0.3,0.5,0.7,0.9\}$ along the 1,000-step training schedule, keep the first temporal latent clean, and run the noised reference latent through the model with its text conditioning. Seed 0 fixes the noise draw, and hooks record the output of all 30 DiT blocks.

The temporal slot mapping follows Wan's $4n+1$ VAE. Slot 0 contains pixel frame 0, while slot $s>0$ contains frames $4s-3$ through $4s$. For each slot, the simulator masks are merged over its pixel frames and max-pooled to the $22\times40$ token grid. If $M_{a,s}$ is the resulting set of tokens for actor $a$, the probe input at block $\ell$ is
\begin{equation}
  \mathbf h^{\ell,\tau}_{a,s}
  = \frac{1}{|M_{a,s}|}
    \sum_{(i,j)\in M_{a,s}} \mathbf h^{\ell,\tau}_{s,i,j}.
  \label{eq:actor-time-pooling}
\end{equation}
A row is omitted when no token is covered. Position is the actor's mean world position over the pixel frames assigned to the slot. Contact is one when any simulator contact in those frames involves the actor. The resulting dataset has one row for every visible example--actor--slot combination.

\begin{table}[H]
  \centering
  \small
  \renewcommand{\arraystretch}{1.13}
  \begin{tabularx}{\linewidth}{@{}p{0.26\linewidth}>{\raggedright\arraybackslash}X@{}}
    \toprule
    Component & Configuration \\
    \midrule
    Backbones & Randomly initialized Wan architecture, base Wan 2.2 TI2V-5B, and the full fine-tuning checkpoint at step 2,500 \\
    Extraction sites & 30 blocks at five schedule fractions, giving 150 cells per backbone \\
    Probe rows & 148,303 training, 19,696 validation, and 18,637 test actor--time rows after visibility filtering \\
    Linear probe & One affine map from 3,072 dimensions to the target \\
    MLP probe & 3,072--64--target with a ReLU hidden layer \\
    Optimization & Adam for 20 epochs, batch size 4,096, learning rate $10^{-3}$, seed 0 \\
    Objectives and scores & MSE and test $R^2$ for position; binary cross-entropy and test ROC-AUC for contact \\
    \bottomrule
  \end{tabularx}
  \caption{\textbf{Probe configuration.} Every layer, schedule fraction, target, probe class, and label condition is fit separately.}
  \label{tab:probe-configuration}
\end{table}

\subsection{Probe Fits and Noise-Resolved Results}

Every cell uses the same example-level split, so rows from one video cannot cross splits. For the shuffled-label control, the training labels are permuted once with the same seed. Validation and test labels remain unchanged. The random backbone retains Wan's VAE, token layout, and architecture but resets the DiT parameters. It separates information already present in the input representation from information organized by pretraining.

\begin{table}[H]
  \centering
  \small
  \setlength{\tabcolsep}{4.2pt}
  \renewcommand{\arraystretch}{1.13}
  \begin{tabular*}{\linewidth}{@{\extracolsep{\fill}}lrrrr@{}}
    \toprule
    & \multicolumn{2}{c}{Position ($R^2$)} & \multicolumn{2}{c}{Contact (AUC)} \\
    \cmidrule(lr){2-3}\cmidrule(lr){4-5}
    Model & Linear & MLP & Linear & MLP \\
    \midrule
    Random init & $0.388\;(-0.007)$ & $0.769\;(-0.011)$ & $0.696\;(0.489)$ & $0.766\;(0.489)$ \\
    Base Wan & $\underline{0.709}\;(-0.006)$ & $\underline{0.878}\;(-0.003)$ & $\underline{0.799}\;(0.506)$ & $\underline{0.816}\;(0.515)$ \\
    Full fine-tuning & $\mathbf{0.785}\;(-0.008)$ & $\mathbf{0.924}\;(-0.006)$ & $\mathbf{0.837}\;(0.505)$ & $\mathbf{0.864}\;(0.509)$ \\
    \bottomrule
  \end{tabular*}
  \caption{\textbf{Probe scores across layers and noise levels.} Each entry is the mean test score over 150 layer--noise cells. The score from a probe trained on shuffled labels is shown in parentheses. Best and second-best trained scores are bold and underlined.}
  \label{tab:probe-score-summary}
\end{table}

The shuffled controls remain close to $R^2=0$ and AUC $=0.5$. Base Wan improves substantially over the random network under a linear readout, and full fine-tuning raises the mean score for both targets and both probe classes. The gap is smaller for the MLP because a nonlinear head can recover much of the spatial information already present in the VAE latent and token layout.

\begin{figure}[H]
  \centering
  \includegraphics[width=\linewidth]{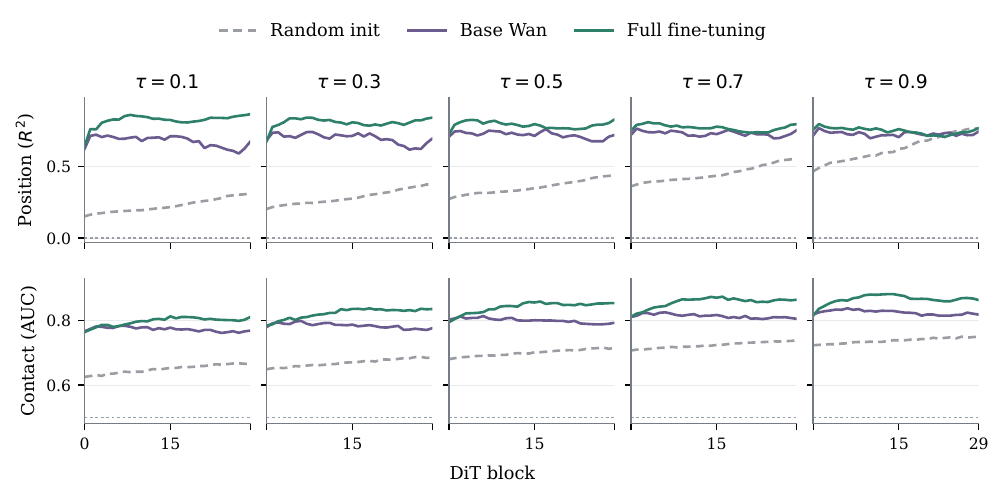}
  \caption{\textbf{Linear probes at every block and schedule fraction.} Each point is one of the 150 fitted cells per model and target. Faint horizontal lines mark chance. The random network approaches the trained models for position at $\tau=0.9$, while the separation is clear at the other schedule positions.}
  \label{fig:probe-linear-by-noise}
\end{figure}

\begin{figure}[H]
  \centering
  \includegraphics[width=\linewidth]{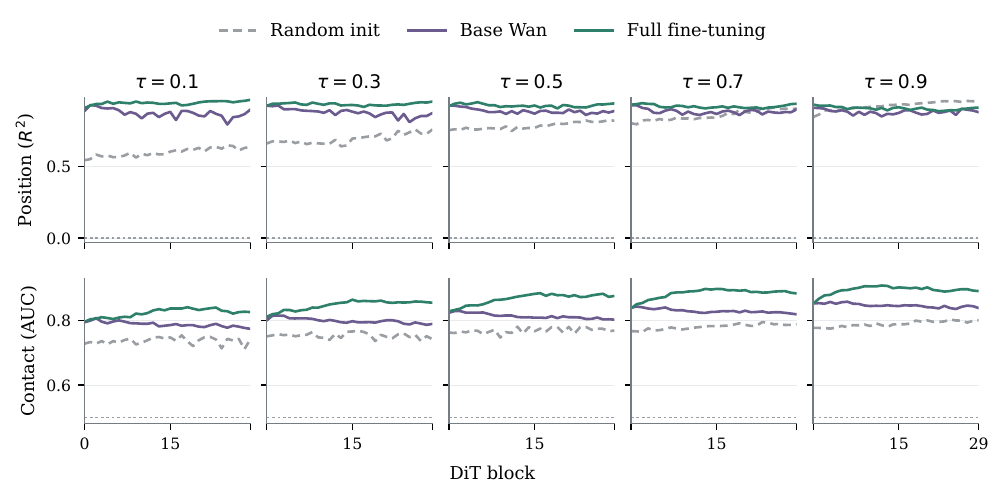}
  \caption{\textbf{MLP probes at every block and schedule fraction.} The nonlinear readout narrows the difference between random and trained networks, especially for position. Contact retains a clearer ordering between the three backbones.}
  \label{fig:probe-mlp-by-noise}
\end{figure}

Additional probe sweeps use simulator quantities aligned to the same actor--time rows. \Cref{tab:probe-state-summary} reports their mean score across all cells and their strongest individual cell. Position remains strong across the sweep, while orientation, angular velocity, and contact normal remain weak. The MLP finds stronger acceleration and velocity scores in a few cells, but these signals do not persist across layers and noise levels.

\begin{table}[H]
  \centering
  \small
  \setlength{\tabcolsep}{4.2pt}
  \renewcommand{\arraystretch}{1.13}
  \begin{tabular*}{\linewidth}{@{\extracolsep{\fill}}lrrrr@{}}
    \toprule
    & \multicolumn{2}{c}{Linear} & \multicolumn{2}{c}{MLP} \\
    \cmidrule(lr){2-3}\cmidrule(lr){4-5}
    Simulator target & Mean & Best cell & Mean & Best cell \\
    \midrule
    Position & 0.709 & 0.772 & 0.879 & 0.926 \\
    Orientation & 0.024 & 0.055 & 0.035 & 0.065 \\
    Linear velocity & -14.958 & -1.497 & -0.010 & 0.278 \\
    Linear acceleration & -0.001 & 0.353 & 0.149 & 0.493 \\
    Angular velocity & -0.095 & -0.023 & -0.002 & 0.001 \\
    Contact normal & 0.096 & 0.141 & 0.049 & 0.138 \\
    \bottomrule
  \end{tabular*}
  \caption{\textbf{Continuous-state probes on base Wan.} Entries are test $R^2$, averaged over the 150 layer--noise cells or taken at the best cell. A negative value is worse than predicting the test-set mean.}
  \label{tab:probe-state-summary}
\end{table}

\subsection{Position-Aligned Interventions}

The intervention sweep uses base Wan at $\tau\in\{0.3,0.5,0.7\}$. A separate position subspace is fitted for each of the 90 block--schedule cells. Within a cell, valid actor--time rows are randomly split 80/20, and INLP repeatedly fits ridge regression with coefficient $0.05$ on the projected training activations. The row space of each regressor is added to the removed basis until a new regressor has training $R^2<0.05$ or 30 iterations have run. Across the saved cells, the resulting position basis has rank 23--50, with mean rank 33.8 out of 3,072.

The rank-matched control is a random orthonormal basis. To construct the variance-matched control, let $\mathbf X_c$ be the centered activation matrix and $\mathbf B_p$ the position basis, giving removed variance
\begin{equation}
  V_p = \frac{\|\mathbf X_c\mathbf B_p\|_F^2}{n-1}.
\end{equation}
Let $\lambda_1\geq\lambda_2\geq\cdots$ be the eigenvalues of the activation covariance, and choose the smallest $K$ for which $\sum_{i=1}^{K}\lambda_i\geq V_p$. Removing a fraction $\alpha$ of the top-$K$ principal-component subspace removes $\alpha(2-\alpha)\sum_{i=1}^{K}\lambda_i$ variance, so we use
\begin{equation}
  \alpha
  = 1 - \sqrt{1-\frac{V_p}{\sum_{i=1}^{K}\lambda_i}}.
  \label{eq:variance-matched-alpha}
\end{equation}
This control changes directions with at least as much activation variance as the position basis, while matching the total amount removed.

For all three variants, the forward hook applies $\widetilde{\mathbf h}=\mathbf h-\alpha(\mathbf h\mathbf B)\mathbf B^\top$ to every residual token at the chosen block. Actor pooling is used to learn the position basis but is not used by the intervention itself. We compare the flow-matching loss with and without the hook under the same noised latent, excluding the clean first temporal slot. Continuing through the remaining blocks gives the full-model effect. Applying Wan's final time-conditioned head directly to the modified state gives the direct-head effect, and the two readouts coincide exactly at block 29.

The intervention evaluates the first 100 training clips after fitting the bases on the full activation corpus. \Cref{tab:intervention-summary} averages over blocks and the three schedule fractions. Under both readouts, the position-aligned intervention is larger than either control for all 100 clip-level averages and for the mean at every block.

\begin{table}[H]
  \centering
  \small
  \setlength{\tabcolsep}{5pt}
  \renewcommand{\arraystretch}{1.13}
  \begin{tabular*}{\linewidth}{@{\extracolsep{\fill}}lcc@{}}
    \toprule
    Removed subspace & Full model $\Delta\mathcal{L}$ & Direct head $\Delta\mathcal{L}$ \\
    \midrule
    Position (INLP) & $0.602\;[0.563,0.644]$ & $1.229\;[1.157,1.304]$ \\
    Variance-matched PC & $0.023\;[0.020,0.026]$ & $0.055\;[0.052,0.059]$ \\
    Random subspace & $0.012\;[0.012,0.012]$ & $0.012\;[0.011,0.012]$ \\
    \bottomrule
  \end{tabular*}
  \caption{\textbf{Aggregate intervention effects.} Each entry is the mean loss increase followed by a bootstrap 95\% confidence interval. We first average each of the 100 clips over 30 blocks and three noise levels, then resample clips 20,000 times.}
  \label{tab:intervention-summary}
\end{table}

\begin{figure}[H]
  \centering
  \includegraphics[width=\linewidth]{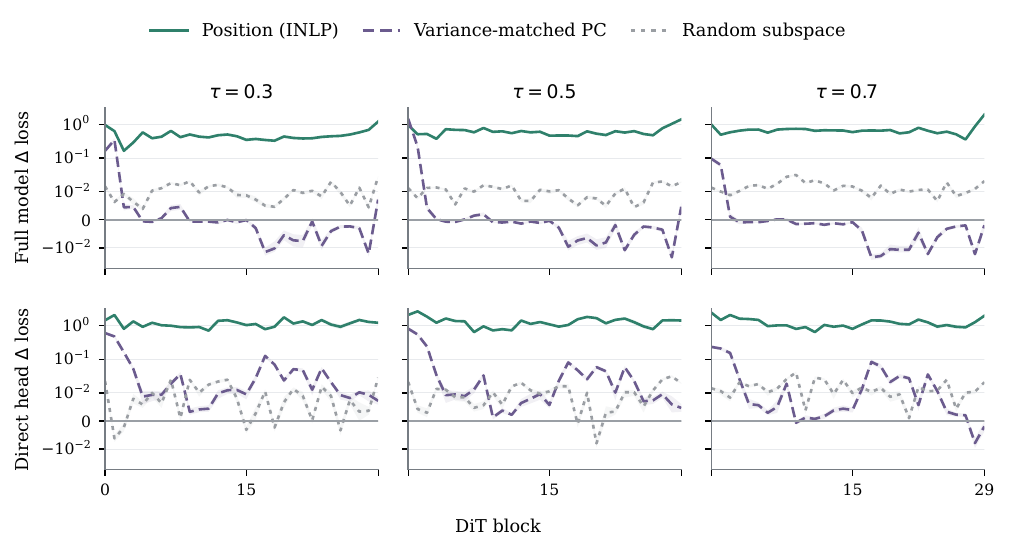}
  \caption{\textbf{Intervention effects by block and schedule fraction.} Lines are means over 100 clips and shading gives bootstrap 95\% confidence intervals. The symmetric-log axis is linear between $-0.01$ and $0.01$, allowing the near-zero controls and the position intervention to remain visible in the same panels.}
  \label{fig:intervention-by-noise}
\end{figure}

\end{document}